\documentclass[letterpaper, 10 pt, conference]{ieeeconf}  

\IEEEoverridecommandlockouts                              

\usepackage{xspace}
\usepackage[dvipsnames]{xcolor}
\usepackage{amsfonts}
\usepackage{graphics}
\graphicspath{{figures/}}
\usepackage{epsfig}
\usepackage{mathptmx}
\usepackage{times}
\usepackage{amsmath}
\usepackage{amssymb}
\usepackage{ulem}
\usepackage{siunitx}
\usepackage{verbatim}
\usepackage{booktabs}
\usepackage{multirow}
\usepackage{makecell}
\usepackage{algorithm}
\usepackage{algpseudocode}
\usepackage{cuted}

\usepackage[font=small,labelfont=bf]{caption}

\makeatletter
\let\origthebibliography\thebibliography
\def\thebibliography#1{%
  \origthebibliography{#1}%
  \scriptsize 
}
\makeatother

\newcommand{\method}{HANDFUL\xspace}
\newcommand{\benchmark}{HANDFUL-Bench\xspace}

\newcommand{\remark}[3]{{\color{#2}[#1: #3]}}
\newcommand{\daniel}[1]{\remark{Daniel}{cyan}{#1}}

\definecolor{Reddish_Purple}{rgb}{0.800, 0.475, 0.655}

\newcommand{\ethan}[1]{\remark{Ethan}{ForestGreen}{#1}}

\definecolor{gold}{rgb}{0.83, 0.68, 0.21}
\definecolor{silver}{rgb}{0.75, 0.75, 0.75}
\definecolor{bronze}{rgb}{0.8, 0.5, 0.2}

\usepackage{color, soul} 

\usepackage[colorlinks=true,citecolor=black,linkcolor=black,urlcolor=black]{hyperref}

\title{\LARGE \bf
\method: Sequential Grasp-Conditioned Dexterous Manipulation with Resource Awareness
}

\author{Ethan Foong$^{*1\S}$, Yunshuang Li$^{*2}$, Hao Jiang$^{2}$, Gaurav S. Sukhatme$^{2}$, Daniel Seita$^{2}$
\thanks{$^*$Equal Contribution.}%
\thanks{$^1$Department of Computer Science at Northwestern University, $^2$Viterbi School of Engineering at the University of Southern California.}%
\thanks{$^\S$This work was initiated through Ethan Foong’s participation in the USC Viterbi Summer Undergraduate Research Experience (SURE) program and continued as a collaboration thereafter.}
}%

\begin{document}
\maketitle
\thispagestyle{empty}
\pagestyle{empty}

\begin{strip}
\vspace{-2cm}
\centering
\captionsetup{type=figure}
\includegraphics[width=1.0\textwidth]{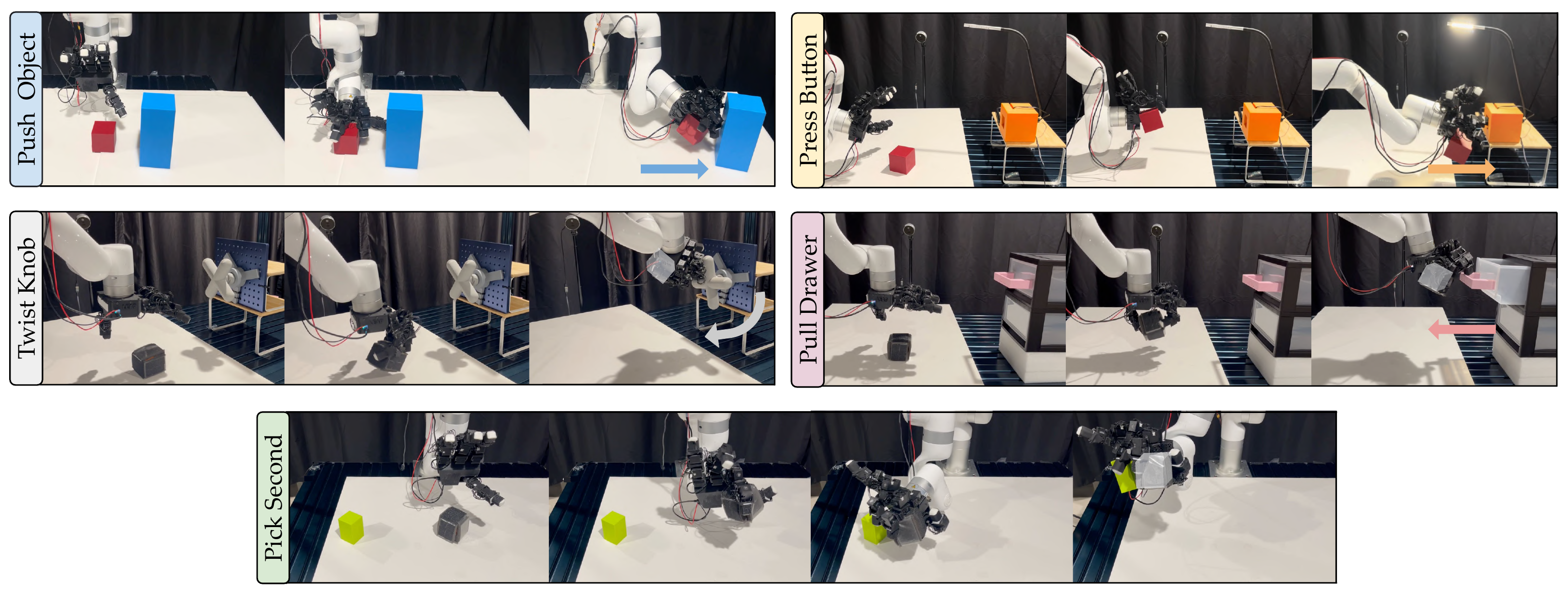}
\vspace{-8pt}
\captionof{figure}{
    \method enables sequential dexterous manipulation by learning resource-aware grasps that preserve specific fingers for downstream subtasks. For each task, the robot first selects an appropriate initial grasp of the object (a block), and then executes a second subtask using the remaining fingers. We show real-world rollouts of \method using a LEAP Hand~\cite{shaw2023leaphand} for tasks that involve pushing (top left), pressing (top right), twisting (middle left), pulling (middle right), and grasping a second object (bottom). 
}
\label{fig:teaser}
\vspace{-0.21cm}
\end{strip}

\begin{abstract}
Dexterous robot hands offer rich opportunities for multifunctional manipulation, where a robot must execute multiple skills in sequence while maintaining control over previously grasped objects. Most prior work in dexterous manipulation focuses on single-object, single-skill tasks. 
In contrast, our insight is that many sequential tasks require resource-aware grasps that conserve fingers for future actions. 
In this paper, we study sequential grasp-conditioned dexterous manipulation, where a robot first grasps an object and then performs a second, distinct manipulation subtask while preserving the initial grasp. We introduce \method, a learning framework that models finger usage as a limited resource and encourages exploration of resource-aware grasps through finger-level contact rewards.
These grasps are subsequently selected for downstream tasks via curriculum-based policy learning. 
We further propose \benchmark, a simulation benchmark that introduces sequential dexterous manipulation tasks across multiple second-subtask objectives, including pushing, pulling, and pressing, under a shared grasp-conditioned setup. Extensive simulation results demonstrate that prioritizing resource-aware grasps improves second-subtask success and robustness compared to a baseline that greedily optimizes the initial grasp before attempting the second subtask. We additionally validate our approach on a real dexterous LEAP hand. Together, this work establishes resource-aware grasp planning as a key principle for multifunctional dexterous manipulation. 
Supplementary material is available on our website: \href{https://handful-dex.github.io//}{\textit{handful-dex.github.io}}.
\end{abstract}

\section{Introduction}

Dexterous multi-fingered hands with high degrees-of-freedom (DoF) offer manipulation capabilities that extend beyond those of parallel-jaw grippers. With many independently actuated fingers, such hands can perform tasks that require great dexterity, such as reorientation of mechanically complex objects~\cite{openai2019solvingrubikscuberobot}. 
However, despite their versatility, much of prior work in dexterous manipulation focuses on isolated skills, such as grasping~\cite{wang2023dexgraspnet,zhang2024dexgraspnet2}, pushing~\cite{li2026learninggeometryaware}, or in-hand rotation~\cite{qi2022hand,qi2023general,yin2023rotatingseeinginhanddexterity}, typically applied to a single object and optimized for immediate task completion.

In contrast, many real-world manipulation scenarios are inherently sequential and multifunctional. A robot organizing a table or workspace must often acquire an object and perform a second task while continuing to hold it. After grasping a container, a robot may need to push a button, pull a drawer, or reach for a second object without relinquishing its initial grasp. These scenarios require the robot to maintain force closure on the held object while simultaneously executing reconfiguration motions with other fingers. Critically, success depends not only on the final grasp stability, but also on how finger contacts are allocated after the first grasp. 
In many cases, grasps optimized purely for force closure or stability occupy fingers and contact regions that are needed for subsequent actions, rendering downstream tasks infeasible. These settings therefore require resource-aware grasps that conserve fingers and contact surfaces for future subtasks. 

In this paper, we propose \method, whole-\textbf{\underline{HAN}}d \textbf{\underline{D}}exterity \textbf{\underline{F}}or seq\textbf{\underline{U}}ential task \textbf{\underline{L}}earning, a framework for sequential dexterous manipulation that treats finger usage as a limited resource to be allocated across subtasks. 
Rather than optimizing grasps for immediate stability, \method learns resource-aware grasping policies that preserve fingers and contact regions for future actions. These grasps are then selected to serve as starting states to downstream subtasks through curriculum-based policy learning, which identifies the grasp modalities that best support subsequent subtasks. 
In addition, we introduce \benchmark, a new simulation benchmark built on ManiSkill~\cite{tao2024maniskill3gpuparallelizedrobotics} with the LEAP Hand~\cite{shaw2023leaphand}. 
The benchmark consists of tasks that share a common first subtask (grasping), while varying the second-subtask objective, such as pushing, pulling, or acquiring a second object. 
By fixing the grasping subtask and varying the second objectives, \benchmark isolates the role of different grasping modalities and enables systematic evaluation of multifunctional dexterous policies. 

To summarize, our main contributions include:
\begin{itemize}
\item A framework for sequential dexterous manipulation that treats finger usage as a limited resource and learns grasp states that preserve resources for future subtasks. 
\item A curriculum-based training approach for efficiently identifying which grasping modalities best support second-subtask policy learning.
\item \benchmark, a simulation benchmark for multi-step dexterous manipulation that highlights the role of different resource-aware grasping modalities in downstream task performance. 
\item Extensive simulation experiments demonstrating improved success rates over stability-focused baselines, along with a retrieval-based execution strategy that enables robust real-world deployment. 
\end{itemize}

\section{Related Work}

\subsection{Dexterous Multi-Object and Multi-Step Manipulation}

Dexterous manipulation is a long-standing area of research~\cite{rus1999inhanddexmanip}.
Broadly, this area attempts to leverage hands with multiple fingers and high DoF to enable greater capabilities. 
Techniques in dexterous manipulation range from analytical (often using variants of force closure analysis)~\cite{liu2022differentiableforceclosure,liu2020deepdifferentiablegraspplanner} to learning-based, such as learning from demonstrations, to perform manipulation tasks using teleoperation~\cite{wang2024dexcap}, or leveraging large-scale datasets for dexterous grasping~\cite{wang2023dexgraspnet,zhang2024dexgraspnet2,xu2023unidexgrasp,lum2024gripmultifingergraspevaluation} or nonprehensile~\cite{li2026learninggeometryaware} maneuvers. 

Within dexterous manipulation, key subareas include \emph{multi-object} and \emph{multi-step} manipulation. 
Multi-object grasping is a well-studied problem~\cite{yamada2015static,yu2001internalforces,yoshikawa2001powergrasp} for which researchers have recently made key advances in analysis~\cite{yao2023exploiting}, algorithms~\cite{he2025sequential,li2024grasp,lu2025graspahandful} and hardware~\cite{eom2024mogrip}. In contrast, we study a general framework for multi-step manipulation which involves controlling multiple objects. Closely-related prior works have studied reinforcement learning approaches for chaining multi-step manipulation policies~\cite{chen2023sequential}. We complement this line of work by studying scenarios where the robot must maintain control of an object during a second subtask. 


\subsection{Simulators and Benchmarks in Robot Manipulation}

Simulators and benchmarks play critical roles in advancing robotic manipulation by enabling rapid, fair, and reproducible comparisons of different algorithms. 
In robot learning for manipulation, researchers utilize simulation frameworks such as IsaacLab~\cite{mittal2025isaaclab} (an updated version of  IsaacGym~\cite{makoviychuk2021isaac}), MuJoCo~\cite{mujoco}, and PyBullet~\cite{coumans2019}. IsaacLab and MuJoCo both provide GPU-acceleration, enabling rapid and large-scale reinforcement learning, and come with robot models~\cite{menagerie2022github} and tasks~\cite{zakka2025mujocoplayground}. 
Closely related to our work, ManiSkill~\cite{tao2024maniskill3gpuparallelizedrobotics} proposes a GPU-parallelized robot simulation framework based on SAPIEN~\cite{xiang2020sapien}. 
ManiSkill also comes with diverse robot manipulation tasks and robot control APIs. This is similar to software benchmarks for testing robot manipulation algorithms, ranging from LIBERO~\cite{liu2023liberobenchmarking}, robosuite~\cite{robosuite2020}, RoboCasa~\cite{nasiriany2024robocasa}, and RLBench~\cite{james2020rlbench}. These are geared for testing generalist manipulation policies. 
Other simulation benchmarks test more specific, yet valuable, manipulation domains, such as deformable object manipulation~\cite{corl2020softgym,seita2021transporters} and dexterous grasping~\cite{bao2023dexartbenchmarking}. 
Our benchmark, \benchmark, is complementary to these works. Built on ManiSkill, it provides the first simulation benchmark to rigorously study sequential and multi-step robotic object manipulation.

\section{Problem Statement and Assumptions}
\label{sec:ps}

We study sequential grasp-conditioned dexterous manipulation in which a ``task'' consists of two ``subtasks,'' ${g^{(1)} \to g^{(2)}}$, where the second must be completed while maintaining the first. 
We assume that the first subtask $g^{(1)}$ is to grasp an object and second subtask $g^{(2)}$ must preserve the initial grasp. The goal is for a robot to autonomously compose and execute both subtasks.
We assume a single-arm robot with a four-fingered hand, such as the LEAP Hand~\cite{shaw2023leaphand} and state-based observations. We assume that the scene contains at least two objects reachable by the robot.


\begin{figure*}[t]
\center
\includegraphics[width=1.00\textwidth]{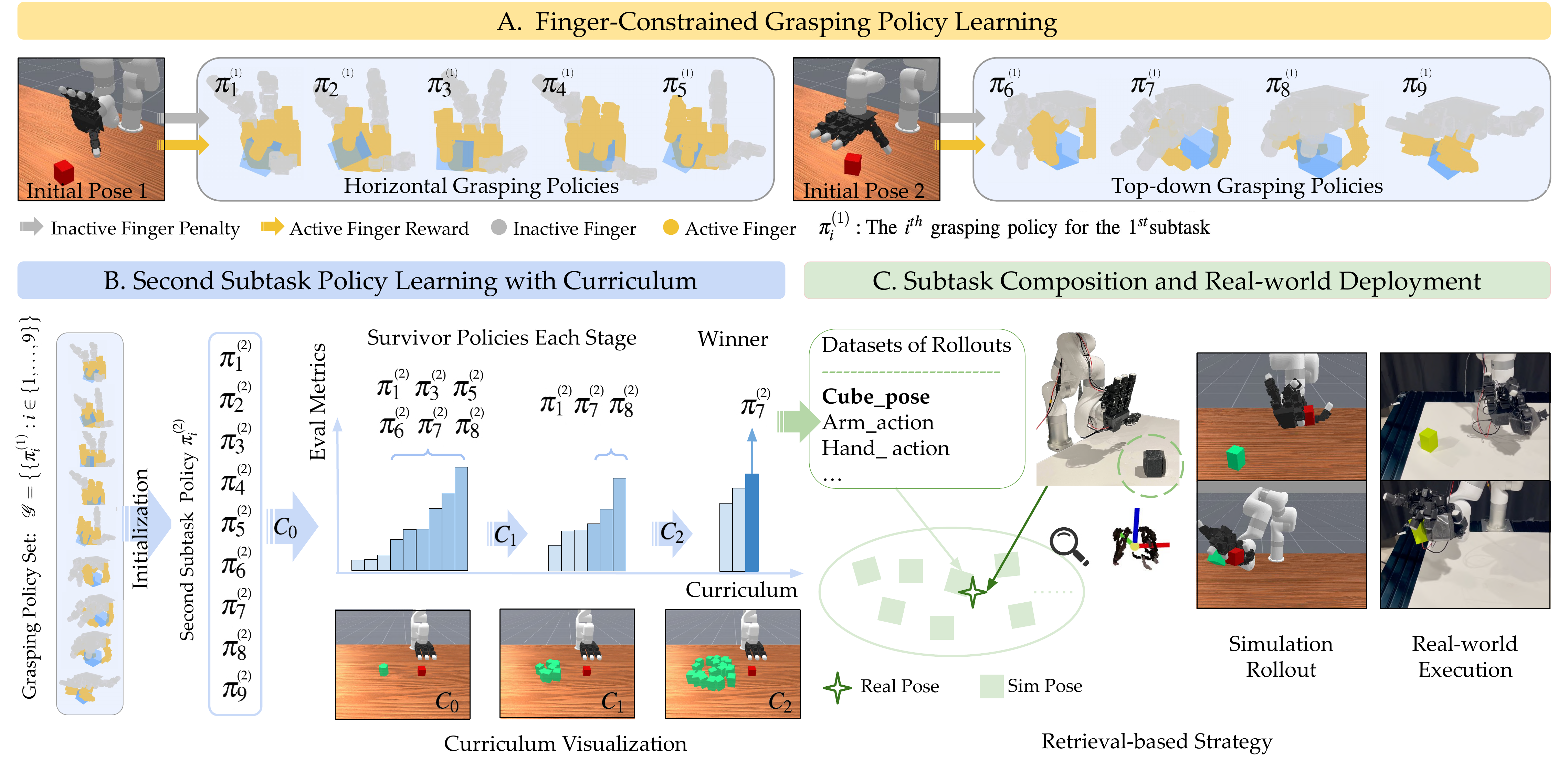}
\caption{
  Overview of \method. 
  \textbf{(A, Sec.~\ref{sec:diverse grasping policies})}: Multiple grasping policies are trained with active and inactive finger constraints to encourage resource-aware grasps that preserve fingers for future manipulation. 
  \textbf{(B, Sec.~\ref{sec: second task policies})}: For each grasp, a second-stage manipulation policy is trained via a multi-stage curriculum, where survivor policies are selected at increasing environment difficulty levels. 
  \textbf{(C, Sec.~\ref{sec:policy chaining})}: Successful policies for grasping and manipulation are composed sequentially. Real-world object poses are matched to simulated rollouts via retrieval-based execution, enabling sim-to-real transfer. 
}
\label{fig:method}
\vspace{-10pt}
\end{figure*}

\section{Method: \method}

\method addresses sequential grasp-conditioned dexterous manipulation by modeling finger usage as a limited resource to be allocated across subtasks. 
We use reinforcement learning to learn two subpolicies that map state-based inputs to continuous actions: a grasping policy $\pi^{(1)}$ that produces resource-aware grasps, followed by a second-subtask policy $\pi^{(2)}$ that completes the task while preserving the initial grasp. 
Our method has three main parts: (1) learning a diverse set of dexterous finger-constrained grasping policies with finger-level contact rewards that conserve unused fingers (Sec.~\ref{sec:diverse grasping policies}); (2) learning dexterous finger-constrained second-subtask policies with curriculum learning to identify grasp modalities that maximize second-subtask success (Sec.~\ref{sec: second task policies}), and (3) composing learned subtasks to deploy them in the real-world via a retrieval-based execution strategy (Sec.~\ref{sec:policy chaining}).

\subsection{Finger-Constrained Grasping Policy Learning}
\label{sec:diverse grasping policies}

The first stage in all benchmarked tasks is grasping. Because the grasping subtask competes with subsequent subtasks for finger allocation, we propose a reward function that explicitly encourages exploration of diverse, low-resource grasping policies. To do this, we define the notion of ``active" and ``inactive" fingers within each subtask. Active fingers specify which fingers interact with the current objective in each subtask, whereas inactive fingers are conserved for the following subtask. For example, in the first grasping subtask, the active fingers are used to grasp and hold the first object.
In the second subtask, the roles reverse: the active fingers from the first subtask stabilize the grasped object, while the previously inactive fingers are used to accomplish the new objective (e.g., twisting a knob or pushing a second block). By incorporating the notion of active and inactive fingers into the reward function, we specify how finger resources are allocated across subtasks within the overall task. 
We define our reward $r_t$ during the grasping subtask at time $t$ as:
\begin{equation}
r_t=w_gr_g+w_vp_v+w_cp_c+w_ar_a+w_{ina}p_{ina}
\end{equation}
where we design the following new reward components:
\begin{align}
r_a &=\frac{1}{|M|}\sum_{i\in{M}}\exp(-\lambda_{a}d_i) & \text{Active Finger Reward} \label{eq:active_reward} \\
p_{ina} &=\operatorname{clip}(-\sum_{j\in J} f_j, -\beta_{ina},0) & \text{Inactive Finger Penalty} \label{eq:inactive_reward}
\end{align}
Here, Eq.~\ref{eq:active_reward} encourages the $M$ active contact points of the active fingers to remain close to the object by increasing the reward as the distance between them and target object $d_i$ decreases. The palm is also included as a contact point in the active contact set. $\lambda_a \in \mathbb{R}$ denotes a scaling factor. Eq.~\ref{eq:inactive_reward} discourages inactive fingers $j\in J$ from touching the object by penalizing their contact forces $f_j$ up to a maximum of $\beta_{ina} \in \mathbb{R}$ to preserve these fingers for the next subtask. In addition, $r_g$ refers to a general grasping reward, adopted from~\cite{SOPE_2024,qin2022dexpointgeneralizablepointcloud,tao2024maniskill3gpuparallelizedrobotics}, with reaching and lifting rewards. Finally, $p_v$ penalizes excessive arm velocity and $p_c$ penalizes collisions with the table. 
The $w_g, w_v, w_c, w_a, w_{ina}$ are scalar weights used to form a weighted sum of the reward components.



To provide diverse starting states for the second subtasks, we train grasping policies using every one- and two-finger combination on the four-finger hand, where selected fingers are designated as active while the remaining fingers are inactive. 
To encourage diverse grasp modalities, we train all grasping policies from two initial hand poses: (1) above the object with the palm facing downward, and (2) level with the object with the palm facing horizontally towards it. This results in \textit{nine} unique and feasible finger configurations (see Fig.~\ref{fig:method}).  
We denote the corresponding set of nine grasping policies as ${\mathcal{G}=\big\{\{\pi_i^{(1)}: i \in \{1, \ldots, 9\} \big\}}$. We train the grasping policies using Soft Actor-Critic (SAC)~\cite{sac}, but \method is compatible with other reinforcement learning algorithms such as Proximal Policy Optimization (PPO)~\cite{ppo}. 

\subsection{Second Subtask Policy Learning with Curriculum}
\label{sec: second task policies}

Starting from a diverse set of grasping policies $\mathcal{G}$, we learn finger-constrained second-subtask policies. We adopt a forward initialization training scheme inspired by Chen~et~al.~\cite{chen2023sequential}, in which each second-subtask policy is trained from terminal states generated by a first-subtask grasping policy $\pi_i^{(1)}$. Different grasps induce distinct hand configurations and contact patterns, which affect the feasibility and performance of downstream tasks. However, training second-subtask policies on top of all nine grasping policies is time-consuming and computationally inefficient. To address this, we introduce a grasp selection strategy combined with curriculum learning to (1) efficiently identify the terminal grasp states best suited for a specific second subtask, and (2) progressively improve second-subtask policy performance.

We construct a curriculum with three stages, ${C_0, C_1, C_2}$, of progressively increasing environment randomization levels. We define a candidate grasp as the terminal states of a grasping policy and the subtask policy initialized from those states. Candidate grasps are ranked in priority order: first by terminal success rate $p_{st}$ of the second-subtask policy, with any-time success rate $p_{sa}$ and average return $p_{ar}$ serving as successive tie-breakers. Training begins in $C_0$ for $k_0$ steps. We then promote six ``survivor'' grasps to $C_1$. After further training, three ``survivor'' grasps are promoted to $C_2$.  
The policy with the highest final success rate is the final ``winner policy'' for the task. Denoting its index as $i^*$, we obtain
$$
\pi^{(1)} = \pi_{i^*}^{(1)}, \quad \pi^{(2)} = \pi_{i^*}^{(2)}
$$
as our final set of policies. See Alg.~\ref{alg:grasp_curriculum} for details.

\begin{algorithm}[t]
\caption{Curriculum-Based Grasp Selection}
\label{alg:grasp_curriculum}
\begin{algorithmic}[1]
\Require $\mathcal{G}$, curriculum $\{C_0,C_1,C_2\}$, steps $\{k_0,k_1,k_2\}$, number of survivor grasps to next stage $m_1=6$, $m_2=3$.

\For{$g_i \in \mathcal{G}$}
    \State Initialize $\pi_i^{(2)}$ from terminal states of $g_i$
    \State Train under $C_0$ for $k_0$ steps
    \State $\Pi \leftarrow \Pi \cup \{\pi_i^{(2)}\}$
\EndFor

\For{$j=1,2$}
    \State Sort $\Pi$ by $p_{st}, p_{sa}, p_{ar}$ \Comment{In priority order}
    \State $\Pi \leftarrow$ select top $m_j$ policies
    \State Continue training under $C_j$ for $k_j$ steps
\EndFor
\State \Return $\arg\max_{\pi \in \Pi} p_1(\pi)$
\end{algorithmic}
\end{algorithm}

The primary reward components for the second-subtasks correspond to their main objective (see Sec.~\ref{sec:benchmark} for the list of second subtasks). However, the active and inactive finger roles reverse between subtasks. Fingers that were active in grasping now hold the object, while previously inactive fingers execute the new manipulation. We therefore retain the active-finger reward (Eq. 2) and inactive-finger penalty (Eq. 3), but apply them to the reversed finger sets. This encourages the original grasping fingers to maintain object control and ensures that the newly active fingers remain free. Finally, any reward terms involving the fingertips in the second subtask are computed using only the active fingertips, allowing the inactive fingertips to prioritize grasp maintenance. 

\subsection{Subtasks Composition and Real-world Deployment}
\label{sec:policy chaining}

To perform the full task, we train the policies in simulation. Then, compose the learned policies by executing $\pi^{(1)}$ to obtain a terminal grasp state, followed by $\pi^{(2)}$ to complete the downstream objective while preserving the grasp. 

While policies are trained in simulation, directly executing them in the real world is difficult due to the sim-to-real gap. To enable safe and efficient real-world deployment, we therefore adopt a retrieval-based strategy built on a dataset of successful simulated rollouts. 
For the best-performing policy of each task in the real world, we collect $N$ trajectories under randomized initial object poses in the first grasping subtask, forming a dataset ${\mathcal{D}=\big\{\tau_k: k \in \{1, \ldots, N\} \big\}}$, where each trajectory ${\tau_k=\{(s_t^{(k)}, a_t^{(k)})\}_{t = 0}^T}$ corresponds to a collision-free execution in simulation. 
During real-world execution, we segment the first object to grasp using SAM2~\cite{ravi2024sam2} and estimate its 3D position ${\mathbf{p}^{\text{real}}\in\mathbb{R}^3}$ from fused point clouds captured by two camera views. We then retrieve the trajectory $\tau_{k^{*}}$ whose initial state best matches the observed pose: 
\begin{equation}
k^{*}=\arg\min_{k} \|\mathbf{p}^{(k)}_0-\mathbf{p}^{\text{real}}\|_2, \quad k \in \{1, \ldots, N\}
\end{equation}
where $\mathbf{p}^{(k)}_0$ denotes the initial first object position in $\tau_k$. We execute the selected trajectory $\tau_{k^{*}}$ in the real world. This retrieval-based strategy improves robustness compared to directly deploying learned RL policies, reducing the sim-to-real gap by grounding execution in pre-validated trajectories.

\begin{figure*}[t]
\center
\includegraphics[width=1.0\textwidth]{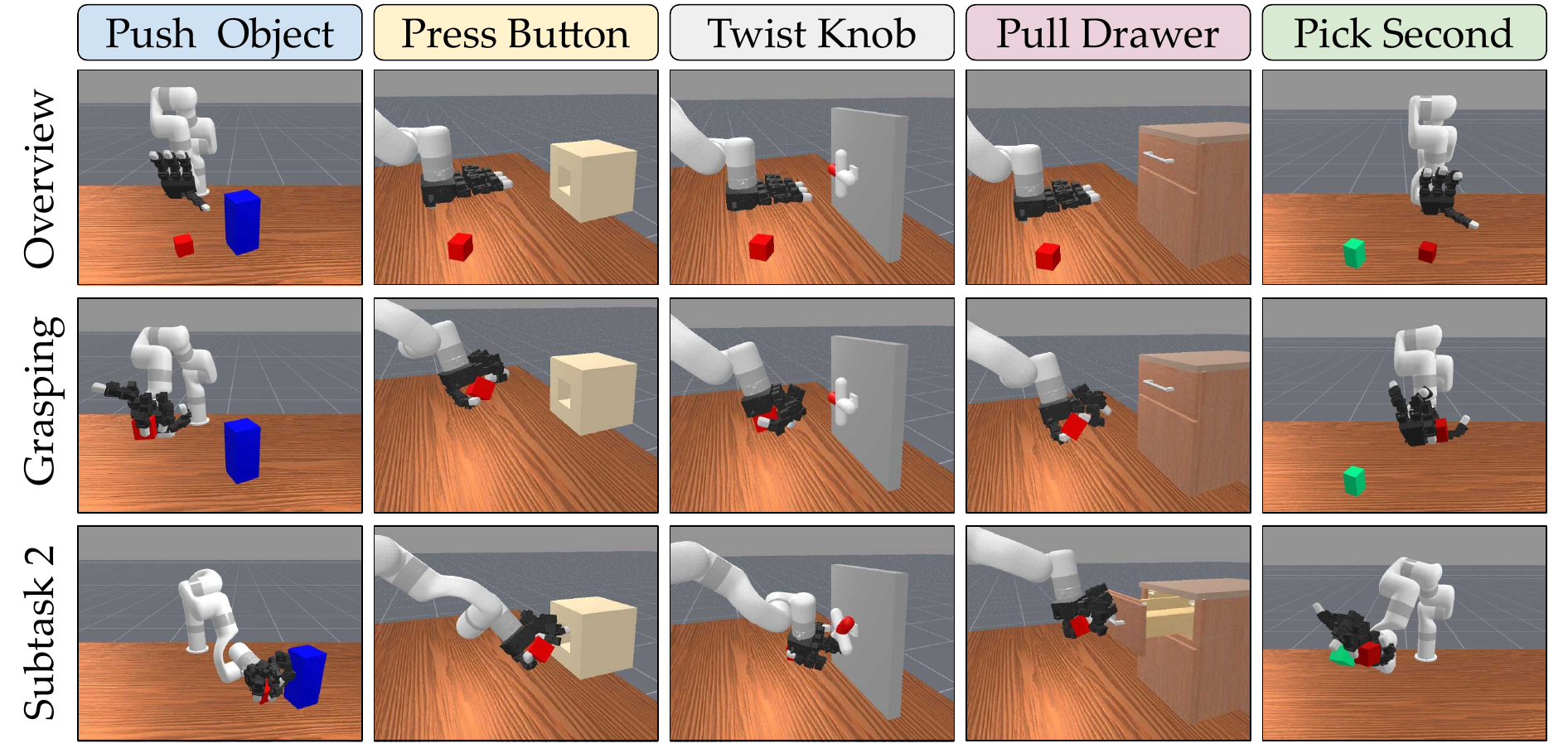}
\caption{
Overview of \benchmark (Sec.~\ref{sec:benchmark}). 
The \underline{\smash{top}} row shows the starting scenarios for the five tasks. 
The \underline{\smash{middle}} row shows the outcome after grasping the block with resource-aware grasping policies. 
The \underline{\smash{bottom}} row shows successful configurations after performing the second subtask while preserving the initial grasp. 
The examples above are test-time executions of \method.
}
\label{fig:sim_benchmark}
\vspace{-10pt}
\end{figure*}

\section{Benchmark: \benchmark}
\label{sec:benchmark}

To accelerate research in sequential dexterous manipulation, we design a benchmark, \benchmark, which will be made open-source. We implement our benchmark in ManiSkill~\cite{tao2024maniskill3gpuparallelizedrobotics}, a robotics simulator with strong community support and GPU acceleration for learning algorithms. 
Each task in \benchmark follows a shared structure. The robot first grasps a target object and then performs one of several second-stage tasks while maintaining the grasp. We consider the following downstream tasks (see Fig.~\ref{fig:sim_benchmark}):
\begin{itemize}
\item \texttt{Push Object}: The robot must push a secondary object, without rotating or tipping it, to a target position.

\item \texttt{Press Button}: The robot must reach and insert its finger into a hole to press a button.

\item \texttt{Twist Knob}: The robot must twist a knob a target number of degrees.

\item \texttt{Pull Drawer}: The robot must pull a drawer open a target distance. 

\item \texttt{Pick Second}: The robot must grasp and lift a second object to a target region.
\end{itemize}


In \benchmark, while the first grasping subtask is shared across all tasks, the second subtasks impose distinct and often competing demands on finger allocation, contact patterns, and space within the hand. Initial grasps optimized solely for force closure can occupy regions of the hand or fingers required for subsequent tasks. As a result, effective manipulation in \benchmark  may require initial grasps that conserve resources by leaving fingers or in-hand spaces free, and in turn, adopt less conventional hand poses to accommodate downstream objectives. Furthermore, grasps that facilitate one second subtask may be suboptimal or infeasible for another, indicating that a diverse set of grasping modalities is necessary for success throughout the task suite. 

Across all environments, we use a 23-dimensional action space consisting of 16 joint commands for the hand and 7 for the arm. Robot control is implemented via a PD delta position controller.
Each subtask provides state observations, which include the arm and hand joint positions and velocities, the palm and fingertip poses, and the arm's wrist position. We also include subtask-specific observations such as object poses, relative pose information (e.g., arm wrist to block), object half sizes (if varying), and articulation joint position values (e.g., knob rotation). We also provide a one hot vector of the current active finger indices in anticipation of policies that can switch grasping or second-subtask execution modalities based on this conditioning. Finally, to encourage stable grasps throughout, episodes run for the full duration of each subtask rather than terminating upon success. Thus, we use terminal success rate $p_{st}$ (whether the task is completed at the final timestep) as our primary evaluation metric.

\section{Simulation Experiments}
\label{sec:sim_exp}

We study the following questions: (1) Is the diversity of grasping policies useful for learning the second subtask? (2) Does curriculum learning increase policy robustness? (3) Do early curriculum stages reliably predict final task performance? (4) Does the selection process retain the best candidate grasps? (5) Does task decomposition help training? 

\subsection{Simulation Experiment Setup}

We use \benchmark with an xArm7 robot arm and a LEAP Hand~\cite{shaw2023leaphand}. 
However, \method is not platform specific and could, in principle, be applied to other dexterous multi-fingered hands (e.g., the Allegro Hand).

\subsection{Methods in Simulation} 
\label{sec: sim methods}

\begin{table*}[t]
\centering
\setlength{\tabcolsep}{6pt}
\renewcommand{\arraystretch}{1.2}

\begin{tabular}{l ccccc}
\toprule
Method
& \texttt{Push Object}
& \texttt{Press Button}
& \texttt{Twist Knob}
& \texttt{Pull Drawer}
& \texttt{Pick Second} \\
\midrule

\method (Ours) 
& $\mathbf{69.90 \pm 5.54}$
& $\mathbf{77.75 \pm 2.15}$
& $\mathbf{61.52 \pm 5.47}$
& $\mathbf{78.94 \pm 1.77}$
& $76.54 \pm 3.63$ \\

w/o Curriculum
& $69.47 \pm 1.18$
& $76.80 \pm 7.32$
& $58.50 \pm 9.48$
& $71.38 \pm 20.16$
& $\mathbf{80.42 \pm 6.72}$ \\

w/o Finger Constraint
& $66.69 \pm 5.66$
& $44.26 \pm 40.58$
& $49.44 \pm 5.73$
& $58.99 \pm 17.36$
& $0.00 \pm 0.00$ \\

Phase-based Reward
& $32.38 \pm 6.36$
& $10.08 \pm 19.23$
& $0.00 \pm 0.00$
& $0.00 \pm 0.00$
& $0.00 \pm 0.00$ \\

\bottomrule
\end{tabular}
\caption{Success rate across tasks in simulation over five seeds. For each task, only the selected grasp is evaluated for both curriculum and non-curriculum training. Here, ``w/o Curriculum'' and ``w/o Finger Constraint'' are ablations of our method (\method) with the identified component removed (see Sec.~\ref{sec: sim methods} for details). For each task, we bold the metric with the highest mean value. 
}
\label{tab:task_method_results_transposed}
\vspace{-10pt}
\end{table*}

We first train nine grasping policies with different hand poses for the first subtask as stated in Sec.~\ref{sec:diverse grasping policies}, achieving an average success rate of $94.67 \pm 2.60$\%, indicating relatively uniform grasping performance. We then initialize nine parallel second-subtask training runs for each task, starting from the successful terminal states generated by each of the grasping policies. These are our candidate grasps. Next, we utilize curriculum-based grasp selection as in Sec.~\ref{sec: second task policies}. The results reported in Table~\ref{tab:task_method_results_transposed} are the performance of ``the winner policy'' evaluated in the second-subtask environments, starting from the matching terminal grasp state. Success is defined as maintaining a stable grasp on the first-subtask object and achieving the goal of the second subtask.

We compare our approach, \method, with the following two ablations and one baseline: 
\begin{itemize}
\item \textbf{w/o Curriculum:} Our method without the curriculum-based training. This directly trains the second-subtask policy once under $C_2$ using the same terminal grasp states as the selected ``winner policy" for each task.
\item \textbf{w/o Finger Constraint:} Our method without learning finger-constrained grasping. All fingers are designated active in both subtasks, with no separation between the fingers responsible for holding the first object and those performing the second-subtask objective.
\item \textbf{Phase-based Reward:} Training the whole task in a unified environment, using two phase-based rewards based on the current timestep of the environment.
\end{itemize}

\subsection{Simulation Results}
\label{sec: sim results}


\subsubsection{Grasping Policy Diversity}
In Table~\ref{tab:task_method_results_transposed}, policies trained with finger constraints (\method and w/o Curriculum) consistently outperform those without (w/o Finger Constraint), with the gap widening for tasks that require greater finger availability, such as \texttt{Press Button}, \texttt{Pull Drawer}, and \texttt{Pick Second}. Notably, w/o Finger Constraint completely fails on \texttt{Pick Second} $(0\%)$, suggesting that without separate finger allocation between subtasks, the policy struggles to free sufficient fingers and in-hand space for dexterous tasks. The high variance on \texttt{Press Button} $(\pm 40.58)$ further reflects this difficulty. This result highlights the importance of promoting and maintaining grasp pose diversity in these grasp-conditioned manipulation tasks.

\subsubsection{Curriculum Learning}
\begin{figure}[t]
\center
\includegraphics[width=0.47\textwidth]{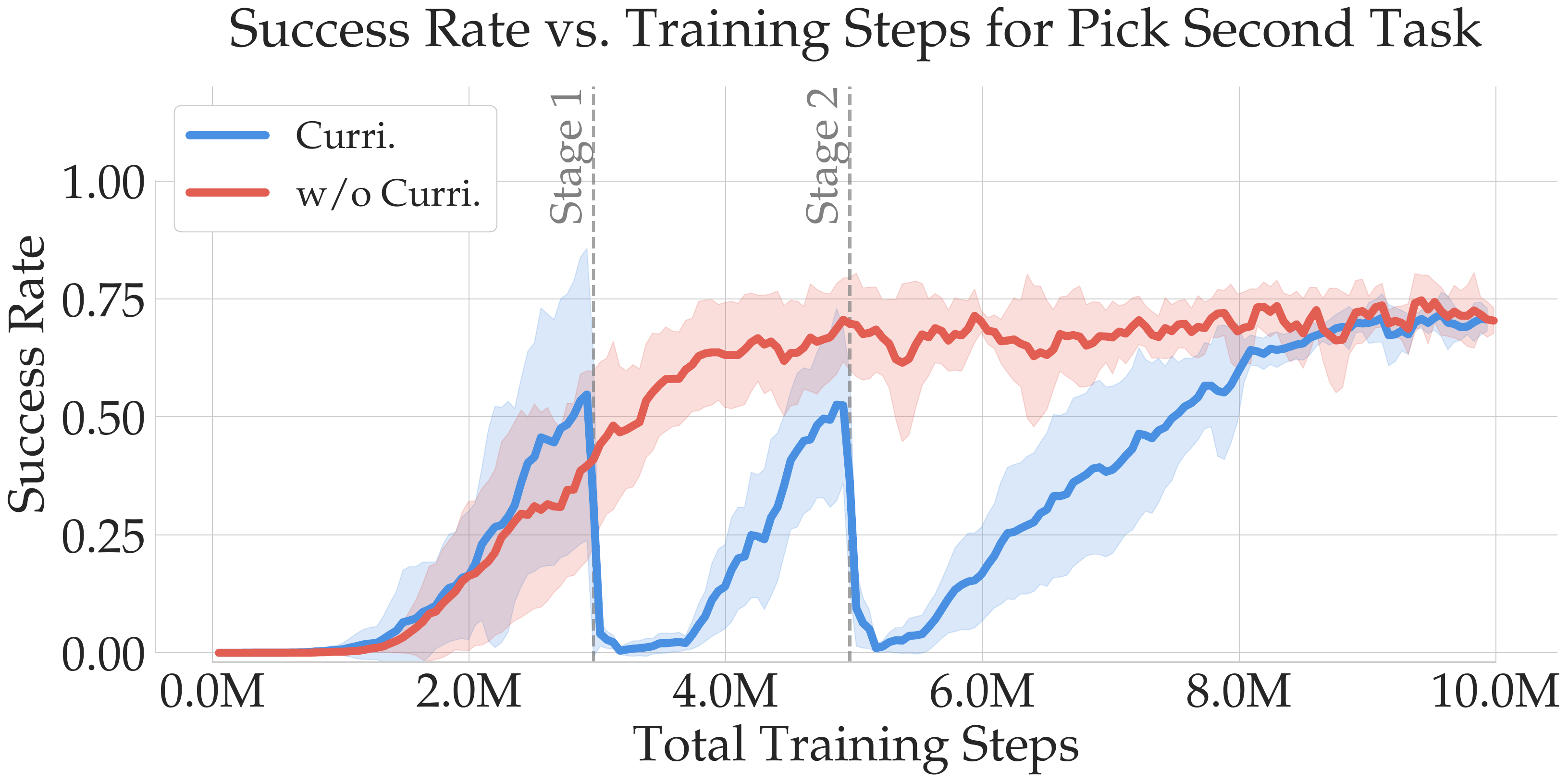}
\caption{
  Performance of \method with curriculum learning vs without curriculum learning over full training with 10 million steps. 
}
\label{fig:curriculum_plot}
\vspace{-10pt}
\end{figure}

See Fig.~\ref{fig:curriculum_plot} for learning curves comparing \method versus w/o Curriculum. \method  achieves comparable success rates (see Table~\ref{tab:task_method_results_transposed}) in all five tasks while reducing total training time by $40\%$. We also observe that curriculum learning may stabilize second-stage policy training, possibly due to early stages of the curriculum providing a more consistent environment to learn successful strategies for each task, though further investigation would be needed to confirm this. Together, these results suggest that the primary contribution of curriculum learning in our setup is its practical benefits to training efficiency and stability.

\subsubsection{Candidate Grasp Selection Validation} 
\label{sec: grasp selection validation}
We investigate whether early curriculum stages reliably predict final task performance and retain the best candidate grasps through a case study on \texttt{Pick Second}. This is a representative task that is a strong test for our selection criteria: training is volatile due to tight constraints on in-hand space and finger placement, and this volatility is consequential. Accidentally eliminating the top 2-3 grasp candidates can cause a larger performance drop off compared to tasks such as \texttt{Push Object}, where all grasps achieve satisfactory performance.

(i) \textit{Predictivity of Early Curriculum Stages:} To assess whether early stages of the curriculum reliably predict final performance, we train all candidate grasps through $C_0, C_1,$ and $C_2$ without selection and record the performance of their second-subtask policies in each stage. Table~\ref{tab:filtering_consistency} shows that the top-performing grasp candidates in $C_2$ consistently perform well in the prior stages, validating that early stage performance is a predictive signal for grasp selection.

(ii) \textit{Selection Stability Across Seeds:} Using the same experiment from (i), we simulate what our selection criteria would select at each stage to evaluate whether it consistently retains the top-performing candidate grasps. Table~\ref{tab:filtering_correctness} shows that the best performing candidate grasp is never prematurely eliminated and the second best is eliminated only once, just before $C_2$. There is some volatility in identifying the third-best grasp, though the grasp candidates selected over it perform only moderately worse in $C_2$ as shown in Table~\ref{tab:filtering_consistency}.


\subsubsection{Task Decomposition}
In Table~\ref{tab:task_method_results_transposed}, all approaches trained with task decomposition significantly outperform our baseline phase-based reward method which fails to solve three tasks on any of the five seeds. Qualitatively, these policies struggle heavily with the shifting reward landscape during rollouts, often prioritizing the second-subtask objectives without learning stable grasps. These results emphasize the importance of task decomposition in multi-step tasks.

\begin{table}[t]
\centering
\resizebox{\columnwidth}{!}{%
\begin{tabular}{cccccccccc}
  \toprule
    Curriculum & $\pi_1^{(2)}$ & $\pi_2^{(2)}$ & $\pi_3^{(2)}$ & $\pi_4^{(2)}$ & $\pi_5^{(2)}$ & $\pi_6^{(2)}$ & $\pi_7^{(2)}$ & $\pi_8^{(2)}$ & $\pi_9^{(2)}$ \\
  \midrule
  $C_0$ & \colorbox{gold}{51.8} & 0.0 & \colorbox{silver}{42.8} & 0.0 & 0.0 & 29.4 & \colorbox{bronze}{36.4} & 18.4 & 1.6\\
  $C_1$ & \colorbox{gold}{65.2} & 0.0 & \colorbox{bronze}{31.8} & 0.0 & 11.8 & 27.4 & \colorbox{silver}{41.2} & 22.5 & 3.6\\
  $C_2$ & \colorbox{gold}{77.8} & 0.0 & 41.8 & 0.1 & 13.2 & 39.3 & \colorbox{silver}{66.0} & \colorbox{bronze}{52.2} & 33.3\\
  \bottomrule
\end{tabular}
}
\caption{Curriculum Predictivity for \texttt{Pick Second}. For each grasp, we report the average maximum training ``success at end'' rate $p_{st}$ over 3 seeds for each stage of the curriculum. Highlighted are the 1st, 2nd, and 3rd best performing grasps in each stage.}
\label{tab:filtering_consistency}
\vspace{-10pt}
\end{table}

\begin{table}[t]
\centering
\resizebox{\columnwidth}{!}{%
\begin{tabular}{cccccccccc}
  \toprule
      Curriculum & $\pi_1^{(2)}$ & $\pi_2^{(2)}$ & $\pi_3^{(2)}$ & $\pi_4^{(2)}$ & $\pi_5^{(2)}$ & $\pi_6^{(2)}$ & $\pi_7^{(2)}$ & $\pi_8^{(2)}$ & $\pi_9^{(2)}$ \\
  \midrule
  $C_0$ & \colorbox{gold}{3} & 3 & 3 & 3 & 3 & 3 & \colorbox{silver}{3} & \colorbox{bronze}{3} & 3 \\
  $C_1$ & \colorbox{gold}{3} & 0 & 3 & 0 & 1 & 3 & \colorbox{silver}{3} & \colorbox{bronze}{2} & 3 \\
  $C_2$ & \colorbox{gold}{3} & 0 & 1 & 0 & 0 & 1 & \colorbox{silver}{2} & \colorbox{bronze}{1} & 1 \\
  \bottomrule
\end{tabular}
}
\caption{Selection Stability Validation for the \texttt{Pick Second} task. For each grasp, we report the number of times (out of 3 seeds) it reached each stage of the curriculum. Highlighted are the 1st, 2nd, and 3rd best performing grasps in the final stage.}
\label{tab:filtering_correctness}
\end{table}


\section{Real-World Experiments}
\label{sec:real_exp}

In this section, we evaluate whether trajectories generated in simulation can reliably transfer to the real world. 

\subsection{Real-World Experiment Setup}
\label{ssec:real_setup}

\begin{figure}[t]
\center
\includegraphics[width=0.49\textwidth]{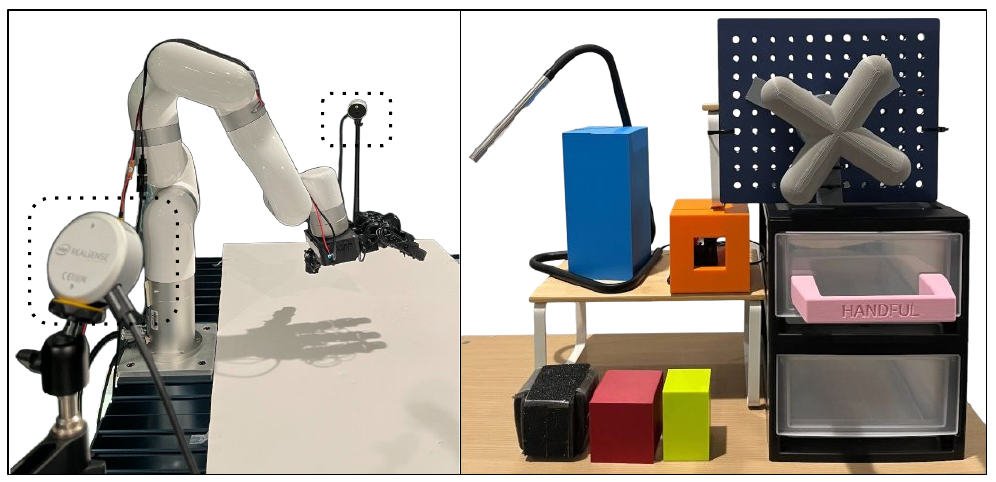}
\caption{
  Left: the real world experiment setup with the xArm7 robot, LEAP hand, and the two Intel L515 cameras indicated with dotted boxes. Right: the different objects we use for experiments. 
}
\label{fig:exp_setup}
\vspace{-10pt}
\end{figure}

As in the simulation setup, all experiments are conducted on a 7-DoF xArm7 robotic manipulator with a four-finger, 16-DoF LEAP Hand~\cite{shaw2023leaphand} (see Fig.~\ref{fig:exp_setup}). 
We use two Intel RealSense L515 RGB-D cameras mounted at complementary viewpoints to estimate the 3D pose of the block. The point clouds from both cameras are fused and segmented using SAM2~\cite{ravi2024sam2} to estimate the block state for retrieval-based execution (Sec.~\ref{sec:policy chaining}).  
To isolate the sequential grasp-conditioned manipulation component of our approach, we fix the the second object and randomize the position of the first block in a 10 by \SI{10}{\centi\meter} region in front of the robot. 

In each evaluation trial, the retrieved trajectory is executed in an open-loop manner until completion. The first subtask succeeds if the robot holds the block until execution ends. The second succeeds if the robot correctly manipulates the second object, regardless of the first block’s state. To analyze performance at different stages, we categorize each trial into four mutually exclusive outcomes. ``Success (Both)" indicates that both Subtask 1 and Subtask 2 are completed successfully within a single trial. ``Fail Subtask 1 Only" denotes trials where Subtask 1 fails but Subtask 2 succeeds. ``Fail Subtask 2 Only" denotes trials where Subtask 2 fails but Subtask 1 succeeds. ``Fail Both" corresponds to trials where neither subtask is successfully completed. These four categories sum to the total number of trials (15 per task).



\subsection{Real-World Experiment Results}
\label{ssec:real_results}


\begin{table}[t]
\centering
\resizebox{\columnwidth}{!}{%
\begin{tabular}{lccccc}
\toprule
\textbf{Tasks} 
& \shortstack{\texttt{Push}\\\texttt{Object}}
& \shortstack{\texttt{Press}\\\texttt{Button}}
& \shortstack{\texttt{Twist}\\\texttt{Knob}}
& \shortstack{\texttt{Pull}\\\texttt{Drawer}}
& \shortstack{\texttt{Pick}\\\texttt{Second}} \\
\midrule
Success (Both) 
& 10 & 8 & 6 & 6 & 4 \\

Fail Subtask 1 Only 
& 4 & 4 & 6 & 6 & 7 \\

Fail Subtask 2 Only 
& 0 & 0 & 0 & 0 & 1 \\

Fail Both 
& 1 & 3 & 3 & 3 & 3 \\
\bottomrule
\end{tabular}
}
\caption{Real-world experiment results. For each task, we report the number of trials (out of 15) categorized by success and failure types. Each column sums to 15 trials. See Sec.~\ref{sec:real_exp} for details.}
\label{tab:real_world_detailed_results}
\end{table}

We deploy our retrieval-based strategy stated in~\ref{sec:policy chaining} with the dataset we collected in simulation, consisting of 50 successful trajectories for each task with randomized block pose in a $\SI{10}{\centi\meter} \times \SI{10}{\centi\meter}$ area in front of the robot.
Over 15 trials per task (Table~\ref{tab:real_world_detailed_results}), \method achieves full two-stage success rates of 66.7\% (\texttt{Push Object}), 53.3\% (\texttt{Press Button}), 40.0\% (\texttt{Twist Knob}), 40.0\% (\texttt{Pull Drawer}), and 26.7\% (\texttt{Pick Second}). Notably, the policy successfully completes at least 40\% of trials on four out of five tasks, despite being trained primarily in simulation.

Across tasks, most failures occur during the second subtask, while isolated second subtask only failures are rare. This indicates that learned grasps are generally stable during grasping and that performance degrades from the increased demands of sequential, multi-functional execution. A representative failure mode is object dropping during the second action, underscoring the challenge of maintaining grasps under changing contact conditions. Overall, these results demonstrate that \method transfers effectively from simulation to the real world and is capable of executing complex, sequential, dexterous manipulation across diverse tasks.

\subsection{Comparison with Demonstration-Based Approach}

\begin{figure}[t]
\center
\includegraphics[width=0.5\textwidth]{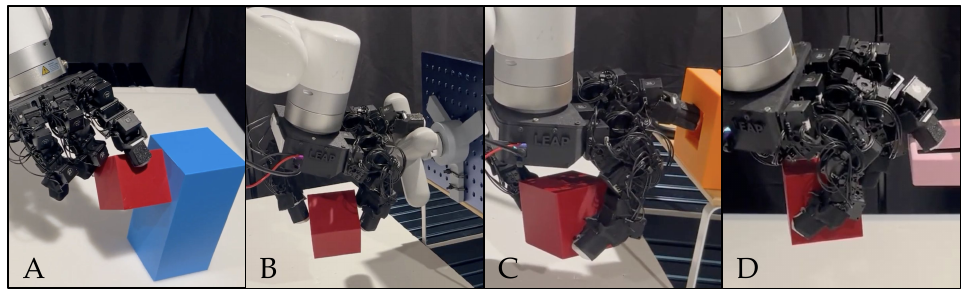}
\caption{
Hand poses in the human teleoperated data. Across tasks, operators predominantly relied on the index finger for the second subtask and used the remaining fingers for grasp stabilization, reflecting a consistent bias and limited pose diversity.
}
\label{fig:human_data}
\vspace{-10pt}
\end{figure}

To complement the results in Sec.~\ref{ssec:real_results}, we evaluate all tasks using 3D Diffusion Policy (DP3)~\cite{Ze2024DP3}, trained on 30 DexCap~\cite{wang2024dexcap} teleoperated demonstrations per task using the segmented block point cloud as input. This comparison highlights the tradeoffs between \method, which relies on reinforcement learning and demonstration-based learning. While fundamental differences in training data prevent a one-to-one comparison, DP3 achieves a 38.6\% success rate over 75 rollouts (15 each task), underperforming our method on four of the five tasks.
We observe systematic differences between the two approaches: teleoperated data rarely leverages palm contact, likely due to teleoperation interface limitations, and exhibits a consistent bias toward using the index finger for the second subtask while reserving other fingers for grasp stabilization, a pattern observed across operators (see Fig.~\ref{fig:human_data} for visualizations). In contrast, \method more explicitly utilizes the palm and produces grasps that fall outside the typical distribution of human demonstrations but better exploit the mechanical capabilities of the robot hand.

\section{Limitations}

Our experiments use a controlled setting, with state-based observations and limited object diversity, which does not fully capture real-world diversity. Additionally, the task suite includes a small set of downstream manipulation objectives chosen to stress finger-level resource constraints; broader task coverage would be needed for generalization to more complex manipulation sequences. While our method is not tied to a specific hand design, we only use the LEAP hand, and future work should test with other robotic hands. Finally, the curriculum relies on manually defined hyperparameters (e.g., survivor counts, duration of curriculum stages). Future work could explore adaptive mechanisms, such as multi-armed bandits, to dynamically select grasps.

\section{Conclusion}

We propose \method, a pipeline for dexterous multi-step manipulation that learns finger-constrained grasping policies and uses curriculum-based grasp selection for downstream subtasks, alongside \benchmark, a simulation benchmark to accelerate future research. 
Our results suggest that in multi-step, multi-object manipulation tasks, effective subtask policy learning may require explicitly accounting for shared physical resources, such as finger availability and in-hand space, that are easy to overlook when learning skills in isolation. We hope that this perspective inspires future work in multi-step and multi-object dexterous hand manipulation.

\section{Acknowledgement}

We are grateful to Yiyang Ling, Sumeet Batra, Shashank Hegde, Darren Chiu, Guangyao Shi and Ellen Novoseller for insightful discussions, and to Zeyu Shangguan and Ruohai Ge for valuable feedback on the manuscript. This work was supported in part by Samsung Research America.

\bibliographystyle{IEEEtranS}
\bibliography{references}

@String { icra    = {IEEE International Conference on Robotics and Automation (ICRA)} }

@String { ieeera  = {IEEE Robotics and Automation Letters (RA-L)} }

@String { ijrr    = {International Journal of Robotics Research (IJRR)} }

@String { iros    = {IEEE/RSJ International Conference on Intelligent Robots and Systems (IROS)} }

@String { isrr    = {International Symposium on Robotics Research (ISRR)} }

@String { rss     = {Robotics: Science and Systems (RSS)} }

@String { tro     = {IEEE Transactions on Robotics} }

@String { neurips = {Neural Information Processing Systems} }

@String { icml    = {International Conference on Machine Learning (ICML)} }

@String { corl    = {Conference on Robot Learning (CoRL)} }

@String { cvpr    = {IEEE Conference on Computer Vision and Pattern Recognition (CVPR)} }

@inproceedings{li2026learninggeometryaware,
  title={{Learning Geometry-Aware Nonprehensile Pushing and Pulling with Dexterous Hands}}, 
  author={Yunshuang Li and Yiyang Ling and Gaurav S. Sukhatme and Daniel Seita},
  booktitle=icra, 
  year={2026},
}

@inproceedings{wang2023dexgraspnet,
  title={Dexgraspnet: A large-scale robotic dexterous grasp dataset for general objects based on simulation},
  author={Wang, Ruicheng and Zhang, Jialiang and Chen, Jiayi and Xu, Yinzhen and Li, Puhao and Liu, Tengyu and Wang, He},
  booktitle=icra,
  year={2023},
}

@inproceedings{seita2021transporters,
  title     = {{Learning to Rearrange Deformable Cables, Fabrics, and Bags with Goal-Conditioned Transporter Networks}},
  author    = {Daniel Seita and Pete Florence and Jonathan Tompson and Erwin Coumans and Vikas Sindhwani and Ken Goldberg and Andy Zeng},
  booktitle = icra,
  Year      = {2021}
}

@inproceedings{yoshikawa2001powergrasp,
  title={Optimization of Power Grasps for Multiple Objects},
  author={T. Yoshikawa and T. Watanabe and M. Daito},
  booktitle=icra,
  year={2001}
}

@inproceedings{he2025sequential,
  title={Sequential Multi-Object Grasping with One Dexterous Hand},
  author={Sicheng He and Zeyu Shangguan and Kuanning Wang and Yongchong Gu and Yuqian Fu and Yanwei Fu and Daniel Seita},
  booktitle=iros,
  year={2025}
}

@inproceedings{mujoco,
  title = {{MuJoCo: A Physics Engine for Model-Based Control}},
  author = {Emanuel Todorov and Tom Erez and Yuval Tassa},
  booktitle = iros,
  year = {2012},
}

@inproceedings{yu2001internalforces,
  title={Computation of Grasp Internal Forces for Stably Grasping Multiple Objects},
  author = {Yong Yu and Kenro Fukuda and Showzow Tsujio},
  booktitle=iros,
  year={2001}
}

@inproceedings{zhang2024dexgraspnet2,
  title={{DexGraspNet 2.0: Learning Generative Dexterous Grasping in Large-scale Synthetic Cluttered Scenes}},
  author={Zhang, Jialiang and Liu, Haoran and Li, Danshi and Yu, XinQiang and Geng, Haoran and Ding, Yufei and Chen, Jiayi and Wang, He},
  booktitle=corl,
  year={2024}
}

@inproceedings{lum2024gripmultifingergraspevaluation,
  title={Get a Grip: Multi-Finger Grasp Evaluation at Scale Enables Robust Sim-to-Real Transfer},
  author={Tyler Ga Wei Lum and Albert H. Li and Preston Culbertson and Krishnan Srinivasan and Aaron D. Ames and Mac Schwager and Jeannette Bohg},
  booktitle=corl,
  year={2024},
}

@inproceedings{chen2023sequential,
  title={Sequential Dexterity: Chaining Dexterous Policies for Long-Horizon Manipulation},
  author={Chen, Yuanpei and Wang, Chen and Fei-Fei, Li and Liu, C Karen},
  booktitle=corl,
  year={2023}
}

@inproceedings{qi2023general,
   title={{General In-Hand Object Rotation with Vision and Touch}},
   author={Qi, Haozhi and Yi, Brent and Suresh, Sudharshan and Lambeta, Mike and Ma, Yi and Calandra, Roberto and Malik, Jitendra},
   booktitle=corl,
   year={2023}
}

@inproceedings{qi2022hand,
   title={{In-Hand Object Rotation via Rapid Motor Adaptation}},
   author={Qi, Haozhi and Kumar, Ashish and Calandra, Roberto and Ma, Yi and Malik, Jitendra},
   booktitle=corl,
   year={2022}
}

@inproceedings{qin2022dexpointgeneralizablepointcloud,
  title={{DexPoint: Generalizable Point Cloud Reinforcement Learning for Sim-to-Real Dexterous Manipulation}}, 
  author={Yuzhe Qin and Binghao Huang and Zhao-Heng Yin and Hao Su and Xiaolong Wang},
  booktitle=corl,
  year={2022},
}

@inproceedings{corl2020softgym,
  title={{SoftGym: Benchmarking Deep Reinforcement Learning for Deformable Object Manipulation}},
  author={Lin, Xingyu and Wang, Yufei and Olkin, Jake and Held, David},
  booktitle=corl,
  year={2020}
}

@inproceedings{zakka2025mujocoplayground,
  title={{MuJoCo Playground}},
  author={Kevin Zakka and Baruch Tabanpour and Qiayuan Liao and Mustafa Haiderbhai and Samuel Holt and Jing Yuan Luo and Arthur Allshire and Erik Frey and Koushil Sreenath and Lueder A. Kahrs and Carmelo Sferrazza and Yuval Tassa and Pieter Abbeel},
  booktitle=rss,
  year={2025},
}

@inproceedings{wang2024dexcap,
  title = {{DexCap: Scalable and Portable Mocap Data Collection System for Dexterous Manipulation}},
  author = {Wang, Chen and Shi, Haochen and Wang, Weizhuo and Zhang, Ruohan and Fei-Fei, Li and Liu, C. Karen},
  booktitle = rss,
  year = {2024}
}

@inproceedings{Ze2024DP3,
  title={{3D Diffusion Policy: Generalizable Visuomotor Policy Learning via Simple 3D Representations}},
  author={Yanjie Ze and Gu Zhang and Kangning Zhang and Chenyuan Hu and Muhan Wang and Huazhe Xu},
  booktitle=rss,
  year={2024}
}

@inproceedings{nasiriany2024robocasa,
  title={{RoboCasa: Large-Scale Simulation of Everyday Tasks for Generalist Robots}},
  author={Soroush Nasiriany and Abhiram Maddukuri and Lance Zhang and Adeet Parikh and Aaron Lo and Abhishek Joshi and Ajay Mandlekar and Yuke Zhu},
  booktitle=rss,
  year={2024}
}

@inproceedings{yin2023rotatingseeinginhanddexterity,
  title={{Rotating without Seeing: Towards In-hand Dexterity through Touch}}, 
  author={Zhao-Heng Yin and Binghao Huang and Yuzhe Qin and Qifeng Chen and Xiaolong Wang},
  booktitle=rss,
  year={2023},
}

@inproceedings{shaw2023leaphand,
  title={LEAP Hand: Low-Cost, Efficient, and Anthropomorphic Hand for Robot Learning},
  author={Shaw, Kenneth and Agarwal, Ananye and Pathak, Deepak},
  booktitle=rss,
  year={2023}
}

@inproceedings{liu2020deepdifferentiablegraspplanner,
  title={Deep Differentiable Grasp Planner for High-DOF Grippers}, 
  author={Min Liu and Zherong Pan and Kai Xu and Kanishka Ganguly and Dinesh Manocha},
  year={2020},
  booktitle=rss,
}

@inproceedings{SOPE_2024,
  title={{Learning to Singulate Objects in Packed Environments using a Dexterous Hand}},
  author={Hao Jiang and Yuhai Wang and Hanyang Zhou and Daniel Seita},
  booktitle = isrr,
  year={2024}
}

@inproceedings{bao2023dexartbenchmarking,
  title={DexArt: Benchmarking Generalizable Dexterous Manipulation with Articulated Objects}, 
  author={Chen Bao and Helin Xu and Yuzhe Qin and Xiaolong Wang},
  booktitle=cvpr,
  year={2023},
}

@inproceedings{xu2023unidexgrasp,
  title={Unidexgrasp: Universal robotic dexterous grasping via learning diverse proposal generation and goal-conditioned policy},
  author={Xu, Yinzhen and Wan, Weikang and Zhang, Jialiang and Liu, Haoran and Shan, Zikang and Shen, Hao and Wang, Ruicheng and Geng, Haoran and Weng, Yijia and Chen, Jiayi and others},
  booktitle=cvpr,
  year={2023}
}

@inproceedings{xiang2020sapien,
  title={{SAPIEN: A SimulAted Part-based Interactive ENvironment}}, 
  author={Fanbo Xiang and Yuzhe Qin and Kaichun Mo and Yikuan Xia and Hao Zhu and Fangchen Liu and Minghua Liu and Hanxiao Jiang and Yifu Yuan and He Wang and Li Yi and Angel X. Chang and Leonidas J. Guibas and Hao Su},
  booktitle=cvpr,
  year={2020},
}

@inproceedings{liu2023liberobenchmarking,
  title={{LIBERO: Benchmarking Knowledge Transfer for Lifelong Robot Learning}}, 
  author={Bo Liu and Yifeng Zhu and Chongkai Gao and Yihao Feng and Qiang Liu and Yuke Zhu and Peter Stone},
  booktitle=neurips,
  year={2023},
}

@inproceedings{sac,
  title = {{Soft Actor-Critic: Off-Policy Maximum Entropy Deep Reinforcement Learning with a Stochastic Actor}},
  author = {Tuomas Haarnoja and Aurick Zhou and Pieter Abbeel and Sergey Levine}, 
  booktitle = icml,
  year = {2018},
}

@inproceedings{lu2025graspahandful,
  title={Grasping a Handful: Sequential Multi-Object Dexterous Grasp Generation},
  author={Lu, Haofei and Dong, Yifei and Weng, Zehang and Pokorny, Florian T. and Lundell, Jens and Kragic, Danica},
  booktitle=ieeera,
  year={2025},
}

@inproceedings{li2024grasp,
  title={Grasp Multiple Objects with One Hand},
  author={Li, Yuyang and Liu, Bo and Geng, Yiran and Li, Puhao and Yang, Yaodong and Zhu, Yixin and Liu, Tengyu and Huang, Siyuan},
  booktitle=ieeera,
  year={2024}
}

@inproceedings{liu2022differentiableforceclosure,
   title={Synthesizing Diverse and Physically Stable Grasps With Arbitrary Hand Structures Using Differentiable Force Closure Estimator},
   author={Liu, Tengyu and Liu, Zeyu and Jiao, Ziyuan and Zhu, Yixin and Zhu, Song-Chun},
   booktitle=ieeera,
   year={2022},
}

@inproceedings{james2020rlbench,
  title={{RLBench: The Robot Learning Benchmark \& Learning Environment}},
  author={James, Stephen and Ma, Zicong and Rovick Arrojo, David and Davison, Andrew J.},
  booktitle=ieeera,
  year={2020}
}

@inproceedings{yao2023exploiting,
  title={Exploiting kinematic redundancy for robotic grasping of multiple objects},
  author={Yao, Kunpeng and Billard, Aude},
  booktitle=tro,
  year={2023},
}

@inproceedings{rus1999inhanddexmanip,
  title={{In-Hand Dexterous Manipulation of Piecewise-Smooth 3-D Objects}},
  author={Daniela Rus},
  booktitle=ijrr,
  year={1999},
}

@article{eom2024mogrip,
  title = {MOGrip: Gripper for multiobject grasping in pick-and-place tasks using translational movements of fingers},
  author = {Jaemin Eom  and Sung Yol Yu  and Woongbae Kim  and Chunghoon Park  and Kristine Yoonseo Lee  and Kyu-Jin Cho},
  journal = {Science Robotics},
  year = {2024}
}

@article{yamada2015static,
  title={Static grasp stability analysis of multiple spatial objects},
  author={Yamada, Takayoshi and Yamamoto, Hidehiko},
  journal={Journal of Control Science and Engineering},
  volume={3},
  year={2015}
}

@misc{menagerie2022github,
  title = {{MuJoCo Menagerie: A collection of high-quality simulation models for MuJoCo}},
  author = {Zakka, Kevin and Tassa, Yuval and {MuJoCo Menagerie Contributors}},
  year = {2022},
}

@article{mittal2025isaaclab,
   title={{Isaac Lab: A GPU-Accelerated Simulation Framework for Multi-Modal Robot Learning}},
   author={Mittal, Mayank and Roth, Pascal and Tigue, James and Richard, Antoine and others},
   journal={arXiv preprint arXiv:2511.04831},
   year={2025},
}

@article{ppo,
  title={{Proximal Policy Optimization Algorithms}},
  author={John Schulman and Filip Wolski and Prafulla Dhariwal and Alec Radford and Oleg Klimov},
  journal={arXiv:1707.06347},
  year={2017}
}

@article{ravi2024sam2,
  title={{SAM 2: Segment Anything in Images and Videos}},
  author={Ravi, Nikhila and Gabeur, Valentin and Hu, Yuan-Ting and Hu, Ronghang and Ryali, Chaitanya and Ma, Tengyu and Khedr, Haitham and R{\"a}dle, Roman and Rolland, Chloe and Gustafson, Laura and Mintun, Eric and Pan, Junting and Alwala, Kalyan Vasudev and Carion, Nicolas and Wu, Chao-Yuan and Girshick, Ross and Doll{\'a}r, Piotr and Feichtenhofer, Christoph},
  journal={arXiv preprint arXiv:2408.00714},
  year={2024}
}

@article{tao2024maniskill3gpuparallelizedrobotics,
  title={{ManiSkill3: GPU Parallelized Robotics Simulation and Rendering for Generalizable Embodied AI}}, 
  author={Stone Tao and Fanbo Xiang and Arth Shukla and Yuzhe Qin and Xander Hinrichsen and Xiaodi Yuan and Chen Bao and Xinsong Lin and Yulin Liu and Tse-kai Chan and Yuan Gao and Xuanlin Li and Tongzhou Mu and Nan Xiao and Arnav Gurha and Zhiao Huang and Roberto Calandra and Rui Chen and Shan Luo and Hao Su},
  journal={arXiv preprint arXiv:2410.00425},
  year={2024},
}

@article{makoviychuk2021isaac,
  title={{Isaac Gym: High Performance GPU-Based Physics Simulation For Robot Learning}}, 
  author={Viktor Makoviychuk and Lukasz Wawrzyniak and Yunrong Guo and Michelle Lu and Kier Storey and Miles Macklin and David Hoeller and Nikita Rudin and Arthur Allshire and Ankur Handa and Gavriel State},
  journal={arXiv preprint arXiv:2108.10470},
  year={2021},
}

@inproceedings{robosuite2020,
  title={{robosuite: A Modular Simulation Framework and Benchmark for Robot Learning}},
  author={Yuke Zhu and Josiah Wong and Ajay Mandlekar and Roberto Mart\'{i}n-Mart\'{i}n and Abhishek Joshi and Soroush Nasiriany and Yifeng Zhu},
  booktitle={arXiv preprint arXiv:2009.12293},
  year={2020}
}

@MISC{coumans2019,
  title =    {{PyBullet, a Python Module for Physics Simulation for Games, Robotics and Machine Learning}},
  author =   {Erwin Coumans and Yunfei Bai},
  year = {2016--2020}
}

@article{openai2019solvingrubikscuberobot,
  title={Solving Rubik's Cube with a Robot Hand}, 
  author={OpenAI and Ilge Akkaya and Marcin Andrychowicz and Maciek Chociej and Mateusz Litwin and Bob McGrew and Arthur Petron and Alex Paino and Matthias Plappert and Glenn Powell and Raphael Ribas and Jonas Schneider and Nikolas Tezak and Jerry Tworek and Peter Welinder and Lilian Weng and Qiming Yuan and Wojciech Zaremba and Lei Zhang},
  journal={arXiv preprint arXiv:1910.07113},
  year={2019},
}

\newpage
\cleardoublepage
\section{APPENDIX}

\subsection{Candidate Grasp Selection Validation on All Tasks}

We extend the case study from Section \ref{sec: grasp selection validation} to all tasks. Table~\ref{tab:unified_predictivity} and Table~\ref{tab:unified_consistency} report curriculum predictivity and selection stability across \texttt{Push Object}, \texttt{Press Button}, \texttt{Twist Knob}, and \texttt{Pull Drawer}.

\begin{table}[htbp]
\centering
\small
\resizebox{\columnwidth}{!}{%

\begin{tabular}{l ccccccccc}
\toprule
Curriculum & $\pi_1$ & $\pi_2$ & $\pi_3$ & $\pi_4$ & $\pi_5$ & $\pi_6$ & $\pi_7$ & $\pi_8$ & $\pi_9$ \\
\midrule
\textit{Push Object} \\
\quad $C_0$ & \colorbox{silver}{80.9} & 10.9 & 51.1 & \colorbox{bronze}{76.2} & 58.3 & 31.4 & 45.7 & \colorbox{gold}{86.7} & 71.4 \\
\quad $C_1$ & \colorbox{silver}{78.2} & 20.2 & 58.2 & \colorbox{bronze}{67.5} & 52.1 & 29.5 & 44.7 & \colorbox{gold}{80.5} & 65.8 \\
\quad $C_2$ & \colorbox{bronze}{58.6} & 28.5 & 21.5 & 46.2 & 52.3 & 43.4 & 54.4 & \colorbox{gold}{75.3} & \colorbox{silver}{62.3} \\
\midrule
\textit{Press Button} \\
\quad $C_0$ & 32.2 & 18.7 & 49.6 & 19.9 & 26.1 & \colorbox{bronze}{74.7} & 73.4 & \colorbox{gold}{84.8} & \colorbox{silver}{76.4} \\
\quad $C_1$ & 18.8 & 3.8 & 42.5 & 25.2 & 57.1 & \colorbox{bronze}{58.0} & \colorbox{silver}{58.3} & \colorbox{gold}{82.5} & 54.0 \\
\quad $C_2$ & 48.9 & 10.2 & 62.2 & 42.8 & 56.7 & \colorbox{bronze}{68.0} & \colorbox{silver}{71.2} & \colorbox{gold}{83.1} & 55.7 \\

\midrule
\textit{Twist Knob} \\
\quad $C_0$ & \colorbox{silver}{30.1} & 0.5 & 14.3 & 0.7 & 4.0 & 4.5 & 14.9 & \colorbox{gold}{57.5} & \colorbox{bronze}{28.8} \\
\quad $C_1$ & \colorbox{silver}{32.7} & 0.7 & 25.3 & 1.2 & 7.0 & 10.5 & 18.9 & \colorbox{gold}{61.3} & \colorbox{bronze}{29.8} \\
\quad $C_2$ & \colorbox{silver}{45.2} & 10.2 & \colorbox{bronze}{38.7} & 4.0 & 15.6 & 16.5 & 28.8 & \colorbox{gold}{63.7} & 38.3 \\
\midrule
\textit{Pull Drawer} \\
\quad $C_0$ & \colorbox{bronze}{48.0} & 0.0 & 8.2 & 1.8 & 0.0 & 0.0 & 19.1 & \colorbox{silver}{59.6} & \colorbox{gold}{65.9} \\
\quad $C_1$ & \colorbox{bronze}{56.0} & 0.0 & 3.1 & 0.0 & 0.0 & 0.0 & 11.6 & \colorbox{gold}{79.8} & \colorbox{silver}{63.3} \\
\quad $C_2$ & \colorbox{silver}{69.3} & 0.0 & 9.2 & 1.8 & 0.0 & 0.0 & 11.7 & \colorbox{gold}{83.2} & \colorbox{bronze}{68.5}\\

\bottomrule
\end{tabular}
}
\caption{\textbf{Early Stage Curriculum Predictivity.} We report the average terminal success rate $p_{st}$ (\%) over 3 seeds for each curriculum stage. This shows how performance on early stages ($C_0, C_1$) correlates with final stage success ($C_2$). Gold, silver, and bronze highlights indicate the top three performing policies per row.}
\label{tab:unified_predictivity}
\end{table}
\begin{table}[t]
\centering
\small
\begin{tabular}{l ccccccccc}
\toprule
Curriculum & $\pi_1$ & $\pi_2$ & $\pi_3$ & $\pi_4$ & $\pi_5$ & $\pi_6$ & $\pi_7$ & $\pi_8$ & $\pi_9$ \\
\midrule
\textit{Push Object} \\
\quad $C_0$ & \colorbox{bronze}{3} & 3 & 3 & 3 & 3 & 3 & 3 & \colorbox{gold}{3} & \colorbox{silver}{3} \\
\quad $C_1$ & \colorbox{bronze}{3} & 0 & 2 & 3 & 2 & 0 & 2 & \colorbox{gold}{3} & \colorbox{silver}{3} \\
\quad $C_2$ & \colorbox{bronze}{3} & 0 & 0 & 2 & 0 & 0 & 0 & \colorbox{gold}{3} & \colorbox{silver}{1} \\
\midrule
\textit{Press Button} \\
\quad $C_0$ & 3 & 3 & 3 & 3 & 3 & \colorbox{bronze}{3} & \colorbox{silver}{3} & \colorbox{gold}{3} & 3 \\
\quad $C_1$ & 2 & 0 & 2 & 1 & 1 & \colorbox{bronze}{3} & \colorbox{silver}{3} & \colorbox{gold}{3} & 3 \\
\quad $C_2$ & 0 & 0 & 2 & 0 & 1 & \colorbox{bronze}{1} & \colorbox{silver}{1} & \colorbox{gold}{3} & 1 \\
\midrule
\textit{Twist Knob} \\
\quad $C_0$ & \colorbox{silver}{3} & 3 & \colorbox{bronze}{3} & 3 & 3 & 3 & 3 & \colorbox{gold}{3} & 3 \\
\quad $C_1$ & \colorbox{silver}{3} & 1 & \colorbox{bronze}{2} & 1 & 2 & 1 & 2 & \colorbox{gold}{3} & 3 \\
\quad $C_2$ & \colorbox{silver}{2} & 0 & \colorbox{bronze}{1} & 0 & 0 & 1 & 1 & \colorbox{gold}{3} & 1 \\
\midrule
\textit{Pull Drawer} \\
\quad $C_0$ & \colorbox{silver}{3} & 3 & 3 & 3 & 3 & 3 & 3 & \colorbox{gold}{3} & \colorbox{bronze}{3} \\
\quad $C_1$ & \colorbox{silver}{3} & 0 & 1 & 1 & 2 & 3 & 2 & \colorbox{gold}{3} & \colorbox{bronze}{3} \\
\quad $C_2$ & \colorbox{silver}{3} & 0 & 0 & 0 & 0 & 0 & 0 & \colorbox{gold}{3} & \colorbox{bronze}{3} \\
\bottomrule
\end{tabular}
\caption{\textbf{Selection Stability across Tasks.} We report the number of seeds (out of 3) that successfully transition through curriculum stages $C_i$ for nine candidate grasping policies ($\pi_1$ to $\pi_9$). Gold, silver, and bronze highlights denote the top three policies by average final stage success rate.}
\label{tab:unified_consistency}
\end{table}

\subsubsection{Predictivity of Early Curriculum Stages} The pattern observed for \texttt{Pick Second} in the main paper holds broadly across tasks. The top-performing candidates in the final curriculum stage consistently rank among the leaders in earlier stages, confirming that early performance is a reliable signal for grasp selection. Early curriculum stages seem more predictive for more difficult tasks (e.g., \texttt{Twist Knob}, \texttt{Pull Drawer}), where there is high variance between grasp types allowing early stages to more clearly reveal separation between strong and weak grasp candidates.

\subsubsection{Selection Stability Across Seeds} Selection stability is similarly strong across tasks. The best-performing candidate grasp is never prematurely eliminated in any seed of any task, and the second and third-best candidates are often retained. As in \texttt{Pick Second}, there is some volatility in selecting these second and third-best grasps, but the impact is minor since the candidates selected over them still tend to perform comparably (e.g., $\pi_9$ in Twist Knob).
\subsubsection{Limitations} Because we train only a single grasping seed per candidate strategy, it is difficult to cleanly attribute each grasp's second-subtask performance to its intrinsic task suitability versus the stability of that particular grasp instance (which may vary with the training seed). In our experiments, candidates such as $\pi_1$ and $\pi_8$ perform strongly across nearly all tasks, which may reflect broad task compatibility, high grasp stability in this seed, or both. Running multiple grasping seeds per strategy and evaluating them on second subtasks would help disentangle these factors and yield more concrete conclusions about which grasps are optimal for each task. Videos of the best grasp strategies per second-subtask are available on our website: https://handful-dex.github.io/.

\begin{table}[t]
\centering
\small
\begin{tabular}{ll}
\toprule
\textbf{Hyperparameter} & \textbf{Value} \\
\midrule
\textit{Training Duration} \\
\quad Grasping stages & 6M steps \\
\quad Second-subtask stages ($C_0$, $C_1$, $C_2$) & 3M, 2M, 5M steps \\
\quad Learning starts ($C_0$ and Grasping) & 51{,}200 steps \\
\quad Learning starts ($C_1$, $C_2$) & 409{,}600 steps \\
\midrule
\textit{Environment} \\
\quad Parallel training environments & 512 \\
\quad Episode Length (Grasping) & 75 \\
\quad Episode Length (All second-subtasks) & 100 \\
\midrule
\textit{SAC} \\
\quad Replay buffer size & 4{,}000{,}000 \\
\quad Batch size & 1{,}024 \\
\quad Training frequency & 2{,}048 steps \\
\quad Update-to-data ratio (UTD) & 0.5 \\
\quad Discount factor $\gamma$ & 0.95 \\
\quad Target network smoothing $\tau$ & 0.005 \\
\quad Policy learning rate & $3 \times 10^{-4}$ \\
\quad Critic learning rate & $3 \times 10^{-4}$ \\
\quad Entropy coefficient $\alpha$ & 0.2 (autotuned) \\
\quad Policy update frequency & 1 \\
\quad Target network update frequency & 1 \\
\bottomrule
\end{tabular}
\caption{\textbf{Hyperparameters of \method.} All values are shared across tasks unless otherwise noted.}
\label{tab:hyperparams}
\end{table}

\subsection{Hyperparameters of SAC}

All experiments use Soft Actor-Critic (SAC)~\cite{sac} with the hyperparameters listed in Table~\ref{tab:hyperparams}. Grasping policies are trained for 6M steps. Second-subtask policies are trained across three curriculum stages $(C_0, C_1, C_2)$ of 3M, 2M, and 5M steps respectively (10M total); in the second and third stages, learning starts after 409,600 steps to allow the carried-over policy to populate the replay buffer with rollouts in the new stage environment before gradient updates begin.

\subsection{Domain Randomization}

To promote robust policies, we apply domain randomization across all tasks per episode. The grasped block is initialized within a 10 by 10 cm square region with a random yaw rotation (intentionally conservative, as the tabletop must simultaneously accommodate all task-relevant objects). Second-subtask object placement follows task-specific distributions. The push block is spawned within a 10 by 10 cm square and the push goal pose within a 40 by 40 cm one. Both are spawned with the same random yaw. The knob is displaced by up to $\pm 15$ cm with $\pm 15^{\circ}$ rotation noise and a uniformly randomized initial joint position in $[-\pi, \pi]$ radians. The button is displaced by up to $\pm 25$ cm with $\pm 15^{\circ}$ rotation noise. The cabinet is placed along a $90^{\circ}$ arc with radius $0.8$ m facing the robot base. It is then perturbed an additional $\pm 5$ cm and  $\pm 15^{\circ}$ from that position and rotation. Finally, the second block in \texttt{Pick Second} is initialized within a 30 by 30 cm square. 

Physical properties are randomized for the grasped block and all task-relevant objects. Mass is scaled by a uniform factor in $[0.5, 1.5]\times$ the nominal value, and surface friction and restitution are sampled uniformly from $[0.1, 0.5]$ and $[0.0, 0.1]$ respectively. The same mass and friction randomization is applied to all robot links. Finally, the grasped block is sampled from a set of five size variants spanning approximately $2.5$--$3.0$ cm per side.

As described briefly in Section \ref{sec: second task policies}, we apply environment randomizations within our second-subtask environments based on curriculum stage. In $C_0$ we'd like our second-subtask policies to focus on finding successful strategies for the subtask so we remove almost all environment randomizations other than the initial position of the hand and grasped object (which come from the end states of our grasping policies). In $C_1$ we introduce position and rotation randomizations at half strength to allow the policies trained in $C_0$ under no randomizations to slowly generalize to more diverse initial conditions. Finally, in $C_2$ we add all randomizations at full strength (including mass, friction, and restitution which are not randomized in $C_0$ and $C_1$) to obtain the final policy. Visualizations of curriculum stage randomization distributions are available on the \href{https://handful-dex.github.io/}{project website}.

\subsection{Subtask Reward and Success Criteria Details}

Throughout sections~\ref{sec:grasp_reward},~\ref{sec:push_reward},~\ref{sec:press_reward},~\ref{sec:knob_reward}, ~\ref{sec:drawer_reward}, and~\ref{sec:two_pick_reward}, $M$ denotes the grasping finger set as defined in Section \ref{sec:diverse grasping policies}, and $J$ denotes the second subtask manipulation finger set (which is all non-grasping fingers). These sets are fixed across all subtasks. As in the main paper, $M$ is active during the grasping subtask and $J$ is active during the second subtasks. One subtlety is that in the grasping subtask only, $M$ includes the palm as an additional contact point to enable palm grasps. We use a similar idea to aid grasping of the second object in \texttt{Pick Second}. We do not find this necessary for initial grasp maintenance or later manipulation in other subtasks.

For the success criteria of all subtasks, two conditions we commonly check are the grasped block being off the table and all of the $M$ grasping fingers being within \SI{0.07}{\meter} of the grasped object's center. This helps us ensure that throughout all the subtasks, the initial grasped object stays grasped by the correct fingers. This is especially important in the second subtasks where the robot hand may accomplish the main objective (like pulling the drawer), but inadvertently drop the grasped object. For simplicity, we combine these conditions and refer to them as whether the object is grasped.

\subsection{Grasping Policy Reward and Success Criteria}
\label{sec:grasp_reward}
In the grasping subtask, success is triggered when the robot lifts the target block within \SI{0.03}{\meter} of a goal location while the block is grasped. 

Following the same reward structure and notation as Section \ref{sec:diverse grasping policies}, the full reward $r_t$ at time $t$ is:
\begin{equation}
\begin{split}
r_t = \; & w_r r_r + w_{gr} r_{gr} + w_l r_l + w_a r_a \\
         & + w_{ina} p_{ina} + w_v p_v + w_c p_c
\end{split}
\end{equation}
where $r_r$, $r_{gr}$, and $r_l$ together constitute the general grasping reward $r_g$ from the main paper:
\begin{align}
r_r &= 1 - \tanh(\lambda_r d_r) & \text{Reach Rew.} \\
r_{gr} &= \frac{1}{|M|}\sum_{i \in M} \mathbf{1}(d_i < \alpha_g) & \text{Grasp Rew.} \\
r_l &= 1 - \tanh(\lambda_l d_l) & \text{Lift Rew.} \\
r_a &= \frac{1}{|M|}\sum_{i \in M} \exp(-\lambda_a d_i) & \text{Active Finger Rew.} \\
p_{ina} &=\operatorname{clip}(-\sum_{j\in J} f_j, -\beta_{ina},0) & \text{Inactive Finger Pen.} \\
p_v &= -\|\mathbf{v}_a\|_2 & \text{Velocity Pen.} \\
p_c &= \operatorname{clip}(-\lambda_c T, -\beta_c, 0) & \text{Contact Pen.}
\end{align}
with hyperparameter definitions in Table~\ref{tab:grasp_reward}. We use weights $w_r=2.0$, $w_{gr}=2.5$, $w_l=40.0$, $w_a=7.5$, $w_{ina}=1.0$, $w_v=1.5$, $w_c=1.0$.

\begin{table}[]
\begin{center}
\begin{tabular}{l l c}
\toprule
\textbf{Symbol} & \textbf{Description} & \textbf{Value} \\
\midrule
$d_r$ & Distance from palm to target block & --- \\
$d_i$ & Distance from contact point $i \in M$ to block  & --- \\
$d_l$ & Distance from block to goal & --- \\
$\sum_{j\in J} f_j$ & Total force of inactive fingers on target block & --- \\
$T$ & Total contact force of hand on table & --- \\
$\mathbf{v}_a$ & Robot joint velocity vector & --- \\
\midrule
$\lambda_r$ & Scaling factor for reach reward & 5.0 \\
$\lambda_a$ & Scaling factor for active finger reward & 20.0 \\
$\lambda_l$ & Scaling factor for lift reward & 5.0 \\
$\lambda_c$ & Scaling factor for collision penalty & 0.05 \\
$\alpha_g$ & Grasp contact threshold & 0.05 m \\
$\beta_{ina}$ & Inactive finger penalty cap & 1.5 \\
$\beta_c$ & Collision penalty cap & 1.0 \\
\bottomrule
\end{tabular}
\caption{Variables and parameters for the grasping subtask reward.}
\label{tab:grasp_reward}
\vspace{-10pt}
\end{center}
\end{table}

\textit{Reward Rationale and Details}. Here we expand on the constituent components of $r_g$ from section \ref{sec:diverse grasping policies}. The \textit{reach} reward provides a positive signal when the palm of the hand is near the block. The \textit{grasp} reward increases with the number of active contact points within a threshold $\alpha_g$ of the block, promoting more secure grasps. The \textit{lift} reward increases as the robot lifts the block toward the goal. The remaining components follow the main paper: $r_a$ incentivizes the $M$ active contacts (palm and fingertips) to remain close to the block, $p_{ina}$ discourages the $J$ inactive fingers from contacting the block to preserve them for the next subtask, $p_v$ penalizes excessive arm velocity, and $p_c$ penalizes hand-table collisions.

\begin{table}[]
\begin{center}
\begin{tabular}{l l c}
\toprule
\textbf{Symbol} & \textbf{Description} & \textbf{Value} \\
\midrule
$d_i$ & Distance from finger $i \in M$ to grasped block & --- \\
$d_r$ & Distance from palm to push block & --- \\
$d_j$ & Distance from pushing finger $j\in J$ to push block & --- \\
$d_p$ & Distance from push block to goal & --- \\
$\theta$ & Rotation error between push block and goal & --- \\
$\sum_{j\in J} f_j$ & Total force of pushing fingers on grasped block & --- \\
$T$ & Total force of hand and grasped block on table & --- \\
$\mathbf{v}_a$ & Robot joint velocity vector & --- \\
\midrule
$\lambda_{gs}$ & Scaling factor for grasp stability reward & 20.0 \\
$\lambda_r$ & Scaling factor for reach reward & 5.0 \\
$\lambda_{pp}$ & Scaling factor for push proximity reward & 5.0 \\
$\lambda_\theta$ & Scaling factor for rotation reward & 5.0 \\
$\lambda_c$ & Scaling factor for soft collision penalty & 0.05 \\
$\beta_{int}$ & Interference penalty cap & 1.0 \\
$\epsilon$ & Collision detection threshold & 0.01 \\
$\beta_c$ & Binary collision penalty magnitude & 4.0 \\
$d_0$ & Initial push block to goal distance (average)& 0.4 m \\
\bottomrule
\end{tabular}
\caption{Variables and parameters for the pushing subtask reward.}
\label{tab:push_reward}
\vspace{-10pt}
\end{center}
\end{table}

\subsection{Pushing Policy Reward}
\label{sec:push_reward}

In the pushing subtask, success is defined as the push block being within \SI{0.03}{\meter} and 0.5 radians of the goal site's position and rotation plus whether the initial object (grasped block) is grasped.

For this reward function, we follow the same reward structure and notation as Section~\ref{sec:grasp_reward}. As this is a second subtask, the set of $M$ grasping fingers is now designated inactive and the set of $J$ pushing fingers is active. The full reward $r_t$ at time $t$ is:
\begin{equation}
\begin{split}
r_t = \; & w_{gs} r_{gs} + w_r r_r + w_{pp} r_{pp} + w_p r_p \\
         & + w_\theta r_\theta + w_{int} p_{int} + w_v p_v + w_c p_c
\end{split}
\end{equation}

which is the weighted sum of eight reward components:
\begin{align}
r_{gs} &= \frac{1}{|M|}\sum_{i\in{M}} \exp(-\lambda_{gs} d_i) & \text{Grasp Stability Rew.} \\
r_r &= 1 - \tanh(\lambda_r d_r) & \text{Reach Rew.} \\
r_{pp} &= \frac{1}{|J|}\sum_{j\in{J}} \exp(-\lambda_{pp} d_j) & \text{Push Proximity Rew.} \\
r_p &= 1 - \frac{d_p}{d_0} & \text{Place Rew.} \\
r_\theta &= \exp(-\lambda_\theta \theta) & \text{Rotation Rew.} \\
p_{int} &=\operatorname{clip}(-\sum_{j\in J} f_j, -\beta_{int},0) & \text{Interference Pen.} \\
p_v &= -\|\mathbf{v}_a\|_2 & \text{Velocity Pen.} \\
p_c &= -\beta_c\,\mathbf{1}(T > \epsilon) - \lambda_c T & \text{Collision Pen.}
\end{align}
with hyperparameter definitions in Table~\ref{tab:push_reward}. We use weights $w_{gs}=7.5$, $w_r=3.0$, $w_{pp}=3.0$, $w_p=20.0$, $w_\theta=5.0$, $w_{int}=1.0$, $w_v=0.1$, $w_c=1.0$.

\textit{Reward Rationale and Details}. $r_{gs}$, $p_{int}$ are the renamed versions of $r_a$ and $p_{ina}$ from the grasping subtask. They motivate grasp maintenance with the grasping fingers and discourage pushing fingers from touching the grasped block to focus on the pushing block respectively. $p_v$ is also conceptually identical to the grasping subtask: penalizing joint velocity. The \textit{reach} reward encourages the hand to move toward the push block. The \textit{push proximity} reward incentivizes the $J$ pushing fingers to remain close to the push block. The \textit{place} reward increases linearly as the push block approaches its goal pose, normalized by the average initial distance $d_0$. The \textit{rotation} reward penalizes angular deviation between the push block and goal orientation, where $\theta$ is the geodesic rotation error in radians. The \textit{collision} penalty slightly differs from the grasping subtask: it adds a binary term $-\beta_c\,\mathbf{1}(T > \epsilon)$ on top of the soft term, and $T$ additionally includes contact forces between the grasped block and table to discourage dropping the grasped block.

\begin{table}[]
\begin{center}
\begin{tabular}{l l c}
\toprule
\textbf{Symbol} & \textbf{Description} & \textbf{Value} \\
\midrule
$d_i$ & Distance from finger $i \in M$ to grasped block & --- \\
$d_j$ & Distance from finger $j \in J$ to button & --- \\
$\sum_{j\in J} f_j$ & Total force of pressing fingers on grasped block & --- \\
$T$ & Total force of hand and grasped block on table & --- \\
$\mathbf{v}_a$ & Robot joint velocity vector & --- \\
\midrule
$\lambda_{gs}$ & Scaling factor for grasp stability reward & 20.0 \\
$\lambda_r$ & Scaling factor for reach reward & 5.0 \\
$\lambda_c$ & Scaling factor for collision penalty & 0.05 \\
$\beta_{int}$ & Interference penalty cap & 1.0 \\
\bottomrule
\end{tabular}
\caption{Variables and parameters for the press button subtask reward.}
\label{tab:press_reward}
\vspace{-10pt}
\end{center}
\end{table}

\subsection{Button Press Policy Reward}
\label{sec:press_reward}
In the pressing subtask, success is only triggered if one of the $J$ pressing fingers is within \SI{0.04}{\meter} of the goal site (simulating a button press), the grasped block is grasped and crucially, if there is no contact between the hand and the button guard/donut.

We follow the same reward structure and notation as Section ~\ref{sec:push_reward}. The $M$ grasping fingers are designated to maintain the initial grasp while one of the $J$ pressing fingers must press the button. The full reward $r_t$ at time $t$ is:
\begin{equation}
r_t = w_{gs} r_{gs} + w_r r_r + w_{int} p_{int} + w_v p_v + w_c p_c
\end{equation}
which is the weighted sum of five reward components:
\begin{align}
r_{gs} &= \frac{1}{|M|}\sum_{i\in{M}} \exp(-\lambda_{gs} d_i) & \text{Grasp Stability Rew.} \\
r_r &= 1 - \tanh(\lambda_r \min_{j \in J} d_j) & \text{Reach Rew.} \\
p_{int} &=\operatorname{clip}(-\sum_{j\in J} f_j, -\beta_{int},0) & \text{Interference Pen.} \\
p_v &= -\|\mathbf{v}_a\|_2 & \text{Velocity Pen.} \\
p_c &= -\lambda_c T & \text{Collision Pen.}
\end{align}
with definitions and values in Table~\ref{tab:press_reward}. Weights are set to $w_r=7.5, w_{gs}=2.0, w_{int}=1.0, w_v=0.1, w_c=1.0$.

\textit{Reward Rationale and Details}. $r_{gs}$, $p_{int}$, and $p_v$ are conceptually identical to the pushing subtask. The \textit{reach} reward uses the minimum distance among inactive fingers $j \in J$ to the button rather than an average, encouraging whichever finger is closest to make contact. $p_c$ retains the form from the grasping subtask, but $T$ additionally includes contact forces between any hand link and the obstacle box, encouraging the robot to avoid collision by precisely pressing the button with its fingertip.

\begin{table}[]
\begin{center}
\begin{tabular}{l l c}
\toprule
\textbf{Symbol} & \textbf{Description} & \textbf{Value} \\
\midrule
$d_i$ & Distance from finger $i \in M$ to grasped block & --- \\
$d_{r}$ & Distance from palm to valve & --- \\
$d_j$ & Distance from twisting finger $j \in J$ to valve & --- \\
$\dot{\theta}_{dir}$ & Valve rotation velocity in target direction & --- \\
$\theta$ & Accumulated valve rotation & --- \\
$\sum_{j\in J} f_j$ & Total force of twisting fingers on grasped block & --- \\
$T$ & Total force of hand/grasped block on obstacles & --- \\
$\mathbf{v}_a$ & Robot joint velocity vector & --- \\
\midrule
$\lambda_{gs}$ & Scaling factor for grasp stability reward & 20.0 \\
$\lambda_r$ & Scaling factor for valve reach reward & 5.0 \\
$\lambda_{fr}$ & Scaling factor for fingertip reach reward & 10.0 \\
$\lambda_{vel}$ & Scaling factor for velocity reward & 5.0 \\
$\theta_{succ}$ & Success rotation threshold & $\frac{4\pi}{3}$ \\
$\lambda_c$ & Scaling factor for collision penalty & 0.05 \\
$\beta_{int}$ & Interference penalty cap & 1.0 \\
\bottomrule
\end{tabular}
\caption{Variables and parameters for the twist knob subtask reward. $T$ includes contact forces between the hand and the wall/table; and between the cube and the wall, table, and knob.}
\label{tab:twist_reward}
\vspace{-10pt}
\end{center}
\end{table}

\subsection{Twist Knob Policy Reward}
\label{sec:knob_reward}
In the twist knob subtask, there is a success if the valve is rotated more than $\frac{4\pi}{3}$ radians in the target direction and the grasped block is grasped. The full reward $r_t$ is defined as:

\begin{equation}
\begin{split}
r_t = \; & w_{gs} r_{gs} + w_r r_r + w_{fr} r_{fr} + w_{vel} r_{vel}\\
         & + w_m r_m + w_{int} p_{int} + w_v p_v + w_c p_c
\end{split}
\end{equation}

where the constituent components are:

\begin{align}
r_{gs} &= \frac{1}{|M|}\sum_{i\in{M}} \exp(-\lambda_{gs} d_i) & \text{Grasp Stability Rew.} \\
r_r &= 1 - \tanh(\lambda_r d_{r}) & \text{Reach Rew.} \\
r_{fr} &= \frac{1}{|J|}\sum_{j \in J} \exp(-\lambda_{fr} d_j) & \text{Fingertip Reach Rew.} \\
r_{vel} &= \tanh(\lambda_{vel} \dot{\theta}_{dir}) & \text{Velocity Rew.} \\
r_m &= \operatorname{clip}(\theta / \theta_{succ}, -1, 1) & \text{Motion Rew.} \\
p_{int} &= \operatorname{clip}(-\sum_{j \in J} f_j, -\beta_{int}, 0) & \text{Interference Pen.} \\
p_v &= -\|\mathbf{v}_a\|_2 & \text{Joint Velocity Pen.} \\
p_c &= \operatorname{clip}(-\lambda_c T, -5.0, 0) & \text{Collision Pen.}
\end{align}

with definitions and values in Table \ref{tab:twist_reward}. We set $w_{gs}=7.5, w_r=3.0, w_{fr}=7.5, w_{vel}=3.0, w_m=10.0, w_{int}=1.0, w_v=0.05, w_c=2.0$.

\textit{Reward Rationale and Details}. The reward structure prioritizes both grasp stability and functional manipulation. $r_{gs}$ provides a dense signal ensuring the newly inactive fingers $M$ remain in contact with the grasped block, while $p_{int}$ prevents the active twisting fingers $J$ from exerting unwanted forces on the grasped object. To guide the manipulation, $r_r$ attracts the palm toward the valve, and $r_{fr}$ provides granular guidance by attracting individual $j \in J$ fingertips to the knob handles. Task progress is captured by $r_{vel}$, which rewards rotation velocity in the target direction $\dot{\theta}_{dir}$, and $r_m$, which measures accumulated rotation $\theta$. The collision penalty $p_c$ captures contact forces $T$ across five object pairs: hand–wall, hand–table, grasped block–wall, grasped block–table, and grasped block–knob. The hand-wall and hand-table penalties help reduce unsafe contact, while the grasped block penalties discourage dropping or dislodging of the block by any of the environment object (in particular, the knob).

\begin{table}[]
\begin{center}
\begin{tabular}{l l c}
\toprule
\textbf{Symbol} & \textbf{Description} & \textbf{Value} \\
\midrule
$d_i$ & Distance from finger $i \in M$ to grasped block & --- \\
$d_{r}$ & Distance from palm to handle & --- \\
$d_j$ & Distance from manipulation finger $j \in J$ to handle & --- \\
$q$ & Joint position of cabinet drawer & --- \\
$q_{target}$ & Target joint position for success & --- \\
$\sum_{j\in J} f_j$ & Total force of manipulation fingers on grasped block & --- \\
$T$ & Total force of hand and grasped block on table & --- \\
$\mathbf{v}_a$ & Robot joint velocity vector & --- \\
\midrule
$\lambda_{gs}$ & Scaling factor for grasp stability reward & 20.0 \\
$\lambda_r$ & Scaling factor for handle reach reward & 5.0 \\
$\lambda_{fr}$ & Scaling factor for fingertip reach reward & 5.0 \\
$\lambda_c$ & Scaling factor for collision penalty & 0.05 \\
$\beta_{int}$ & Interference penalty cap & 1.0 \\
\bottomrule
\end{tabular}
\caption{Variables and parameters for the pull drawer subtask reward.}
\label{tab:cabinet_reward}
\vspace{-10pt}
\end{center}
\end{table}

\subsection{Pull Drawer Policy Reward}
\label{sec:drawer_reward}
In the pull drawer subtask, success is triggered if the drawer is opened more than $q_{target}$, the drawer handle is static, and the initial object is grasped. The full reward $r_t$ is:
\begin{equation}
\begin{split}
r_t = \; & w_{gs} r_{gs} + w_r r_r + w_{fr} r_{fr} + w_o r_o \\
         & + w_{int} p_{int} + w_v p_v + w_c p_c
\end{split}
\end{equation}
where the constituent components are:
\begin{align}
r_{gs} &= \frac{1}{|M|}\sum_{i \in M} \exp(-\lambda_{gs} d_i) & \text{Grasp Stability Rew.} \\
r_r &= 1 - \tanh(\lambda_r d_{r}) & \text{Handle Reach Rew.} \\
r_{fr} &= \frac{1}{|J|}\sum_{j \in J} (1 - \tanh(\lambda_{fr} d_j)) & \text{Finger Reach Rew.} \\
r_o &= q / q_{target} & \text{Open Rew.} \\
p_{int} &= \operatorname{clip}(-\sum_{j \in J} f_j, -\beta_{int}, 0) & \text{Interference Pen.} \\
p_v &= -\|\mathbf{v}_a\|_2 & \text{Joint Velocity Pen.} \\
p_c &= -\lambda_c T & \text{Collision Pen.}
\end{align}
with definitions and values in Table~\ref{tab:cabinet_reward}. We set $w_{gs}=5.0$, $w_r=1.0$, $w_{fr}=5.0$, $w_o=15.0$, $w_{int}=1.0$, $w_v=0.1$, $w_c=3.0$.

\textit{Reward Rationale and Details}. $r_{gs}$, $p_{int}$, $p_v$, and $p_c$ are conceptually identical to prior subtasks. $r_r$ attracts the palm toward the handle, and $r_{fr}$ guides the $J$ fingertips toward it. $r_o$ measures normalized joint progress toward the target joint position $q_{target}$. To avoid reward competition, once the joint begins opening, $r_r$ is overridden to its maximum value, and once sufficiently open, both $r_r$ and $r_{fr}$ are similarly overridden, allowing the policy to focus entirely on maintaining the initial grasp.

\begin{table}[]
\begin{center}
\begin{tabular}{l l c}
\toprule
\textbf{Symbol} & \textbf{Description} & \textbf{Value} \\
\midrule
$d_i$ & Distance from finger $i \in M$ to grasped block & --- \\
$d_{r}$ & Distance from palm to second block & --- \\
$d_j$ & Distance from contact point $j$ to second block & --- \\
$d_p$ & Distance from second block to goal & --- \\
$h$ & Current height of second block & --- \\
$h_0$ & Initial height of second block & --- \\
$g_M$ & Fraction of $M$ fingers grasping grasp block & --- \\
$\sum_{j\in J} f_j$ & Total force of $J$ fingers on grasped block & --- \\
$T$ & Total force of robot and grasped block on table & --- \\
$\mathbf{v}_a$ & Robot joint velocity vector & --- \\
\midrule
$\lambda_{gs}$ & Scaling factor for grasp stability reward & 20.0 \\
$\lambda_r$ & Scaling factor for reach reward & 5.0 \\
$\lambda_{fd}$ & Scaling factor for finger distance reward & 20.0 \\
$\lambda_p$ & Scaling factor for place reward & 5.0 \\
$\alpha_g$ & First block grasp contact threshold & 0.07 m \\
$\alpha_p$ & Second block grasp contact threshold & 0.05 m \\
$h_{max}$ & Maximum lift bonus height & 0.1 m \\
$\lambda_c$ & Scaling factor for collision penalty & 0.05 \\
$\beta_{int}$ & Interference penalty cap & 1.0 \\
\bottomrule
\end{tabular}
\caption{Variables and parameters for the two-pick subtask reward.}
\label{tab:two_pick_reward}
\vspace{-10pt}
\end{center}
\end{table}

\subsection{Two-Pick Policy Reward}
\label{sec:two_pick_reward}
In the two-pick subtask, the robot is successful if the second block is lifted within \SI{0.03}{\meter} of the goal site and the initial block is grasped. The full reward $r_t$ is:
\begin{equation}
\begin{split}
r_t = \; & w_{gs} r_{gs} + w_r r_r + w_{fd} r_{fd} + w_{sg} r_{sg} \\
         & + w_h r_h + w_p r_p + w_{int} p_{int} + w_v p_v + w_c p_c
\end{split}
\end{equation}
where the constituent components are:
\begin{align}
r_{gs} &= \frac{1}{|M|}\sum_{i \in M} \exp(-\lambda_{gs} d_i) & \text{Grasp Stability Rew.} \\
r_r &= 1 - \tanh(\lambda_r d_{r}) & \text{Reach Rew.} \\
r_{fd} &= \frac{\sum_{j \in J \cup \{p\}}\exp(-\lambda_{fd} d_j)}{|J| + 2} & \text{Finger Dist.\ Rew.} \\
r_{sg} &= \frac{1}{|J|}\sum_{j \in J} \mathbf{1}(d_j < \alpha_p) & \text{Second Grasp Rew.} \\
r_h &= \operatorname{clip}\!\left(\frac{h - h_0}{h_{max}}, 0, 1\right) \cdot g_M & \text{Height Rew.} \\
r_p &= 1 - \tanh(\lambda_p \, d_p) & \text{Place Rew.} \\
p_{int} &= \operatorname{clip}(-\sum_{j \in J} f_j, -\beta_{int}, 0) & \text{Interference Pen.} \\
p_v &= -\|\mathbf{v}_a\|_2 & \text{Joint Velocity Pen.} \\
p_c &= -\lambda_c T & \text{Collision Pen.}
\end{align}

where $g_M = \frac{1}{|M|}\sum_{i \in M}\mathbf{1}(d_i < \alpha_g)$ is the initial grasped block's grasp fraction. With definitions and values in Table~\ref{tab:two_pick_reward}. We set $w_{gs}=7.5$, $w_r=3.0$, $w_{fd}=7.5$, $w_{sg}=2.5$, $w_h=20.0$, $w_p=30.0$, $w_{int}=1.0$, $w_v=0.05$, $w_c=0.2$.

\textit{Reward Rationale and Details}. $r_{gs}$, $p_{int}$, $p_v$, and $p_c$ are conceptually identical to prior subtasks. $r_r$ draws the palm toward the second block. $r_{fd}$ incentivizes the $J$ fingertips to remain close to the second block. For this term only, we also include the palm as a contact point in $J$ with the palm weighted by a factor of 2 to enable palm + fingertip grasps like our initial grasping stage. $r_{sg}$ provides a binary signal for how many of the $J$ fingertips are within threshold $\alpha_p$ of the block, promoting a secure grasp before lifting. The height reward $r_h$ increases as the second block is lifted above its initial height $h_0$, but is gated by $g_M$ to encourage the policy to maintain the original grasp while lifting. $r_p$ increases as the second block approaches its goal.

\subsection{Sim-to-Real Experiments Details}

When executing rollouts in simulation (results discussed in Sec.~\ref {sec:real_exp}), we randomize the block position within a 10 by \SI{10}{\centi\meter} region in front of the robot, matching the distribution of the data collected during simulation. As mentioned in Sec.~\ref{sec:policy chaining}, we then roll out the learned policies by first executing $\pi^{(1)}$ to obtain a terminal grasp state, followed by $\pi^{(2)}$ to complete the downstream objective while preserving the grasp. For each successful, collision-free rollout, we save the 3D grasped block pose and the full sequence of 23-DoF joint position and velocity observations. To replay the trajectory in the real world, we infer the absolute joint position commands by applying the sum of the joint pose observation and the joint velocity observation scaled by the control time step, $\Delta t$. Because the converged policies are highly robust in simulation, data collection sample efficiency is high; generating 50 successful, collision-free trajectories typically required running fewer than 70 simulated episodes.

\begin{table}[t]
\centering
\resizebox{\columnwidth}{!}{%
\begin{tabular}{lccccccccc}
  \toprule
  Grasping & $\pi_1^{(2)}$ & $\pi_2^{(2)}$ & $\pi_3^{(2)}$ & $\pi_4^{(2)}$ & $\pi_5^{(2)}$ & $\pi_6^{(2)}$ & $\pi_7^{(2)}$ & $\pi_8^{(2)}$ & $\pi_9^{(2)}$ \\
  \midrule
    Sim              & 99.3 & 95.7 & 95.6 & 93.7 & 97.4 & 92.6 & 91.5 & 91.9 & 94.3 \\
  Real - Black & 20 & 80 & 50 & 70 & 50 & 70 & 30 & 50 & 20 \\
  Real - Red   & 0  & 70 & 60 & 80 & 70 & 40 & 10 & 0  & 0  \\
  Real - Avg                 & 10 & 75 & 55 & 75 & 60 & 55 & 20 & 25 & 10 \\
  \bottomrule
\end{tabular}
}
\caption{Success rates of nine grasping policies in both simulation and real-world settings. Evaluations were conducted using two different blocks (see Fig.~\ref{fig:exp_setup}) over a \SI{10}{\centi\meter} by  \SI{10}{\centi\meter} region in the real world for 10 trials each.}
\label{tab:sim2real_grasping}
\end{table}

\begin{figure}[t]
\center
\includegraphics[width=0.47\textwidth]{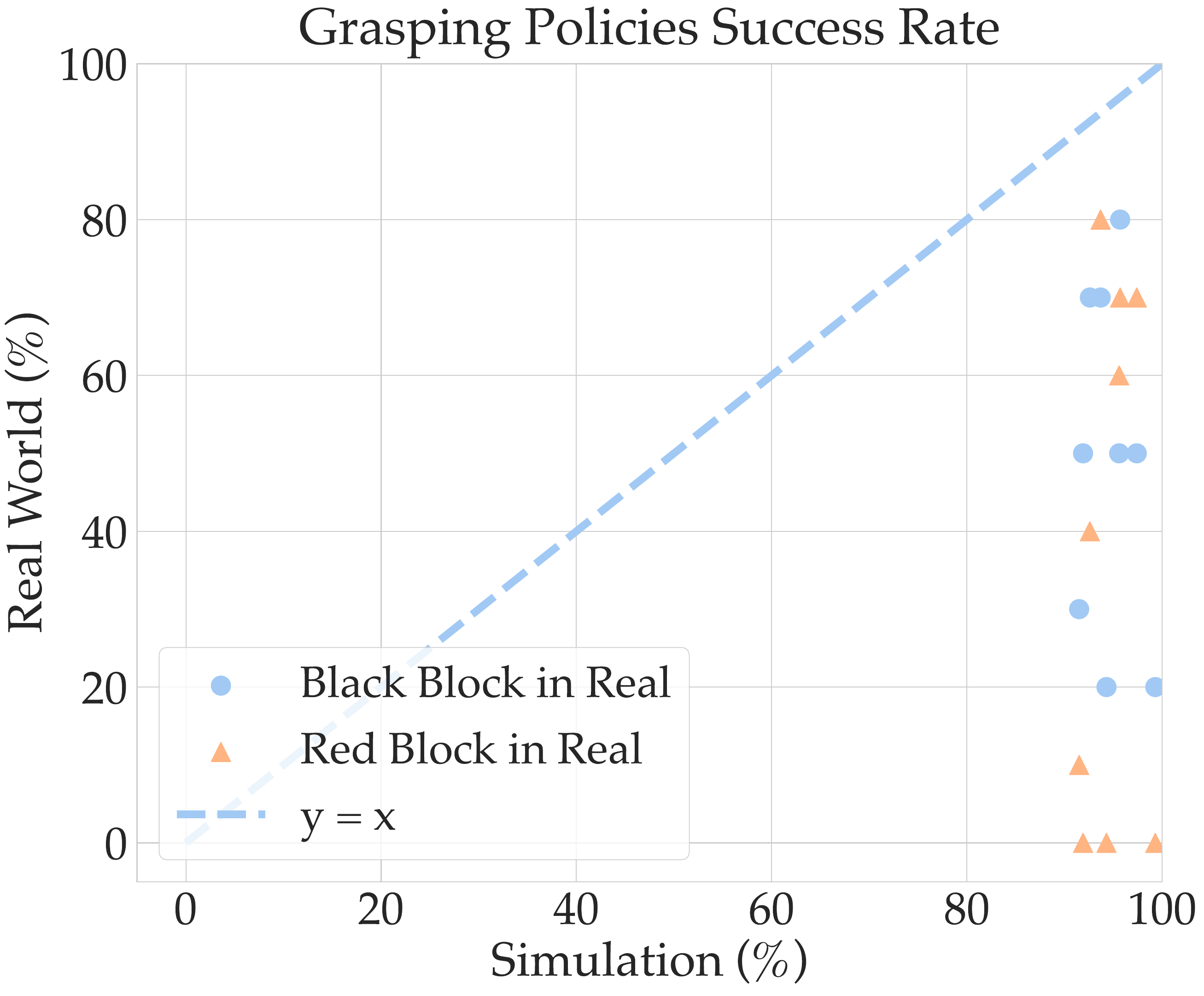}
\caption{
Success rates of the nine grasping policies in simulation versus real-world settings. Each point represents a policy evaluated on either the black or red block. The dashed line $y=x$ denotes the ideal case with no sim-to-real gap, where success rates in simulation and real world are identical. While in reality, simulation performance remains consistently high, real-world results exhibit substantial variability, revealing a notable sim-to-real gap across different grasping policies.
}
\label{fig:distribution_shift}
\vspace{-10pt}
\end{figure}

Before executing the full trajectories of the two subtasks, we first evaluated the nine grasping policies in the real world. The blocks (see Fig.~\ref{fig:exp_setup}) were placed uniformly across the testing region, and each policy was tested under the same conditions. The black block features a slightly deformable surface with higher friction, while both the black and red blocks are approximately 7 by \SI{7}{\centi\meter} cube. Table~\ref{tab:sim2real_grasping} summarizes the success rates in both simulation and real-world settings, and Fig.~\ref{fig:distribution_shift} visualizes these results. We observe a clear distribution shift in success rates across different grasping policies. In simulation, all nine policies achieve consistently high performance with low variance (on average $94.7 \pm 2.5$\%). In contrast, real-world performance varies significantly across policies.

We hypothesize two main reasons for this discrepancy:
(1) More challenging grasping policies rely heavily on real-time adjustments for successful execution. When deployed in an open-loop replay setting, the absence of feedback can significantly degrade performance.
(2) Grasping policies that establish more stable contact exhibit a smaller sim-to-real gap. In particular, policies that utilize the palm ($\pi_1^{(2)}$, $\pi_2^{(2)}$, $\pi_3^{(2)}$, $\pi_4^{(2)}$, $\pi_5^{(2)}$) are less affected, as the larger and more stable contact surface provides greater robustness compared to fingertip-based grasps. This observation supports our design choice of explicitly incorporating palm-based grasping. Therefore, based on the above observations, in real-world experiments (see Sec.~\ref{ssec:real_results}), we execute trajectories whose corresponding first subtask grasping policies both reach the final stage, i.e., $C_2$ of curriculum training and achieve the highest success rates in the preceding evaluations.

\end{document}